\documentclass[11pt]{article}
\usepackage[T1]{fontenc}
\usepackage[a4paper, scale=0.8]{geometry}
\usepackage{setspace}
\usepackage{microtype}
\usepackage{tocloft}
\usepackage{graphicx}
\usepackage{float}
\usepackage{caption}
\usepackage{subcaption}
\usepackage{tabularx}
\usepackage{multicol}
\usepackage{physics}
\usepackage{siunitx}
\usepackage{xcolor}
\usepackage{booktabs, makecell} 
\usepackage{threeparttable}
\definecolor{lightgreen}{RGB}{0.5 1 0.5} 

\usepackage{hyperref}
\usepackage[sort&compress]{natbib}

\usepackage{amsmath}
\usepackage{amssymb}
\usepackage{mathtools}
\usepackage{amsthm}
\usepackage{tikz-cd}
\usepackage{newtxtext}
\usepackage[varvw,subscriptcorrection]{newtxmath}
\usepackage{algorithm,algorithmic}
\newcommand{\bc}{\mathbf c}

\usepackage[capitalize,noabbrev]{cleveref}

\theoremstyle{plain}
\newtheorem{theorem}{Theorem}[section]

\theoremstyle{definition}

\theoremstyle{remark}

\crefname{figure}{Fig.}{Figs.}
\crefname{table}{Table}{Tables}
\crefname{equation}{Eq.}{Eqs.}
\crefname{subsection}{Subsection}{Subsections}

\AtBeginDocument{\RenewCommandCopy\pqty\qty}
\AtBeginDocument{\RenewCommandCopy\qty\SI}
\DeclareSIUnit{\us}{\text{\textmu}s}

\newcommand\bm{\mathbf}
\usepackage[textsize=tiny]{todonotes}

\title{{ReBaNO: Reduced Basis Neural Operator Mitigating Generalization Gaps and Achieving Discretization Invariance}}

\author{
Haolan Zheng\thanks{Department of Mathematics, University of Massachusetts Dartmouth, North Dartmouth, MA 02747, USA. Email: {\tt{hzheng1@umassd.edu}}. Research is supported by National Science Foundation grant DMS-2208277. The code of ReBaNO is available at \href{https://github.com/haolanzheng/rebano}{https://github.com/haolanzheng/rebano}.} 
\and 
Yanlai Chen\thanks{Department of Mathematics, University of Massachusetts Dartmouth, North Dartmouth, MA 02747, USA. Email: {\tt{yanlai.chen@umassd.edu}}. Research is supported in part by National Science Foundation grant DMS-2208277 and by Air Force Office of Scientific Research grant FA9550-25-1-0181.}
\and 
Jiequn Han\thanks{Flatiron Institute, New York, NY 10010, USA. Email: {\tt{jhan@flatironinstitute.org }}.}
\and
Yue Yu\thanks{Department of Mathematics, Lehigh University, Bethlehem, PA 18015, USA. Email: {\tt{yuy214@lehigh.edu}}. Research is supported in part by National Institute of Health grant 1R01GM157589-01.}}

\date{}

\begin{document}

\maketitle

\begin{abstract}
We propose a novel data-lean operator learning algorithm, the Reduced Basis Neural Operator (ReBaNO), to solve a group of PDEs with multiple distinct inputs. Inspired by the Reduced Basis Method and the recently introduced Generative Pre-Trained Physics-Informed Neural Networks, ReBaNO relies on a mathematically rigorous greedy algorithm to build its network structure offline adaptively from the ground up. Knowledge distillation via task-specific activation function allows ReBaNO to have a compact architecture requiring minimal computational cost online while embedding physics. In comparison to state-of-the-art operator learning algorithms such as PCA-Net, DeepONet, FNO, and CNO, numerical results demonstrate that ReBaNO significantly outperforms them in terms of eliminating/shrinking the generalization gap for both in- and out-of-distribution tests and being the only operator learning algorithm achieving strict discretization invariance. 
\end{abstract}

\section{Introduction}\label{sec:intro}
With the advent of deep neural networks, deep learning has achieved enormous success in various fields of science and engineering, especially in solving Partial Differential Equations (PDEs)~\cite{lu2021learning,li2020multipole,li2021fourier,kovachki2023neural}.  In particular, in the problems where repeated and computationally expensive simulations are demanded, deep neural networks are competitive as a fast and reliable solver. As an emerging approach, the paradigm of operator learning has been shown to be computationally efficient compared to traditional numerical methods. Unlike classical deep learning applications that require approximations to the mappings between finite-dimensional vector spaces (for example, from image pixellation or text embeddings), operator learning models aim to infer mappings between infinite-dimensional function spaces. When solving a group of PDEs with distinct inputs, conventionally we need a fine discretization to solve each PDE accurately using certain computational methods. Nevertheless, repeated simulations are computationally prohibitive when dealing with complex systems, especially nonlinear problems. \par 

Recently, Neural Operators (NOs) ~\cite{kovachki2021universal,lu2021learning,li2021fourier,kovachki2023neural,liPhysicsInformedNeuralOperator2024a, wenUFNOAnEnhancedFourier2022, raonicConvolutionalNeuralOperators2023a} were proposed as data-driven deep learning framework for learning operators, and they provide a promising path to alleviate the prohibitive computational cost from repeatedly solving PDEs.  With NOs, one can train a model using labeled data from PDE solutions of multiple inputs, and then employ the model to provide an efficient prediction on new and unseen inputs. Based on the universal approximation theorem~\cite{chenUniversalApproximationNonlinear1995}, neural operators provide discretization convergent approximations to ground truth operators from the data, which outperform the finite-dimensional operators given by traditional deep neural networks. The convergence and error bounds are also guaranteed under some assumptions~\cite{kovachki2021universal,deng2022approximation,lanthaler2023operator,you2022learning}. However, there are three limitations of data-driven neural operators. First, a neural operator requests a large number of annotated input-output pairs for training. To generate sufficiently large training datasets, it usually involves computationally expensive numerical simulations or costly experiments. Second, pure data-driven models can not guarantee  fundamental physics laws, such as the conservation laws. The physics knowledge needs to be implicitly inferred from data, leading to poor generalizability to unseen inputs, in particular, Out-of-Distribution (OOD) inputs~\cite{hoopCostAccuracyTradeOffOperator2022a,mcgreivy2024weak}. Third, neural operators struggle in super/sub-resolution tests when they are trained in the low-resolution regime, meaning that the neural operator does not learn a truly mesh-free approximator to the target operator~\cite{lu2022comprehensive}. To alleviate the first and second limitations, Physics-Informed Neural Operators (PINOs)~\cite{goswami2023physics,liPhysicsInformedNeuralOperator2024a} were introduced, which attempt to embed physical constraints in the loss function. However, PINOs still face difficulties from highly non-convex optimization, long training time and sensitivity to the choice of loss weights. 

To address these issues, we propose the \textbf{Reduced Basis Neural Operator (ReBaNO)}, a data-lean reduced basis-driven operator learning algorithm inspired by the Physics-Informed Neural Network (PINN) \cite{raissi2019physics} and the Generative Pre-Trained Physics-Informed Neural Network (GPT-PINN) \cite{chen2024gpt}. ReBaNO leverages a reduced basis approach. By drawing from the success of PINNs in approximating individual PDE solutions, our method features a similar design of an \textit{offline-online} decomposition. During the offline stage, ReBaNo uses a small number of representative full-order PINN solutions to build up a reduced space during the offline stage. On the online stage, given a new instance of input, the corresponding solution can be obtained efficiently by fine-tuning a lightweight network with a single hidden layer, where each unit, conceptually acting as a `neuron', encapsulates a precomputed full PINN. Compared to existing neural operator methods, our approach has two features. First, it requires \textit{no} high-fidelity training data. Second, while it significantly reduces generalization errors for OOD predictions, while maintaining a competitive accuracy for In-Distribution (ID) predictions. Furthermore, ReBaNO intrinsically adheres to a mesh-free paradigm by integrating mesh-independent PINNs as functional building blocks.

Our main contributions at present include the following.
\begin{enumerate}
    \item we introduce a physics-informed operator learning framework based on a mathematically rigorous reduced basis approach, which adaptively constructs a low-dimensional surrogate space from full-order PINN solutions to improve generalization while maintaining efficiency and accuracy; 
    \item we propose an offline-online decomposition paradigm, eliminating the need for training data and enabling efficient physics-informed inference through a minimal network structure;
    \item we demonstrate that ReBaNO consistently outperforms state-of-the-art operator learning methods in terms of reducing the generalization gaps in OOD tests and being the only one achieving strict discretizations invariance.
\end{enumerate}
The remainder of this paper is structured as follows. First, in \cref{sec:background} we briefly review operator learning models, reduced basis method (RBM), and PINNs. Next, we present problem setting and build a unified framework of operator learning models in \cref{sec:unified-ol-framework}. In \cref{sec:rebano}, we entail the design of ReBaNO, which is empirically validated in \cref{sec:results}, where we compare the performance of ReBaNO with other models on three benchmark problems and showcase the better generalizability and superior mesh invaraiance of ReBaNO. Finally, we conclude this paper in \cref{sec:conclusion}.

\section{Background {and Related Work}}\label{sec:background}

Predicting complex physical responses is ubiquitous in many scientific and engineering applications   \citep{ghaboussi1998autoprogressive,ghaboussi1991knowledge,carleo2019machine,e2021machine,karniadakis2021physics,zhang2018deep,cai2022physics,pfau2020ab,he2021manifold,besnard2006finite,jafarzadehPeridynamicNeuralOperators2024,liuDeepNeuralOperator2024}. Traditionally, the governing law is given as a PDE, and numerical methods, such as finite difference method (FDM) and finite element method (FEM), are developed to discretize the domain into a grid/basis and approximate the PDE solutions on this grid/basis \citep{leveque2007finite,brenner2008mathematical,maday2002reduced}. Unavoidably, since these methods usually rely on a highly fine grid, the generation of grid/basis is time-consuming and may impose truncation errors. 
To accelerate the PDE solving procedure, reduced basis method was introduced, characterized by the greedy algorithm and the efficiency of its offline-online decomposition. Moreover, neural networks, which have demonstrated revolutionary success in modern artificial intelligence tasks, have also been shown to be powerful in solving scientific problems \cite{lu2021learning,e2021machine,kovachki2023neural,raissi2019physics}. In particular, PINNs and operator learning paradigm pave two distinct ways towards scientific machine learning. 

This section is devoted to a concise overview of operator learning models in \cref{subsec:bg-oplearn}, a short introduction to PINN and its variational form in \cref{subsec:bg-pinn}, and then a brief review of RBM in \cref{subsec:bg-rbm}.

\subsection{Neural Operators}\label{subsec:bg-oplearn}

Unlike classical neural network architectures for vector-to-vector mappings, neural operators seek to approximate the mappings between infinite-dimensional Banach spaces as applications of machine learning to PDEs \citep{anandkumarNeuralOperatorGraph2020,liu2024domain,li2020multipole,li2021fourier,you2022nonlocal,ongIntegralAutoencoderNetwork2022,cao2021choose,lu2021learning,goswami2023physics, gupta2021multiwavelet,han2023equivariant}. As the emerging successful paradigm, most of NOs are data-driven deep learning models such as DeepONet~\cite{lu2021learning}, PCA-Net~\cite{bhattacharyaModelReductionNeural2021}, Fourier Neural Operator (FNO)~\cite{li2021fourier} and its variants~\cite{liPhysicsInformedNeuralOperator2024a, wenUFNOAnEnhancedFourier2022, tranFactorizedFourierNeural2022}, and Convolutional Neural Operator (CNO)~\cite{raonicConvolutionalNeuralOperators2023a}. In addition, transformer-based methods, such as Galerkin Transformer~\cite{cao2021choose}, Transolver \cite{wuTransolverFastTransformer2024a,luoTransolverAccurateNeural2025a}, take advantage of attention mechanism to capture physical correlations and handle general geometrics. These neural operators are often employed to approximate the mapping between spatial and/or spatio-temporal function pairs. Compared with classical neural networks, the NOs are more favorable due to their resolution independence, convergence guarantee, and fast inference. More specifically, NOs can be evaluated on any given discretization rather than the mesh on which they are trained~\citep{li2021fourier,zhengalias}. And the convergence of NOs to the target operator and the error estimate are proved given certain assumptions based on the universal approximation theorem for operators~\cite{chenUniversalApproximationNonlinear1995,kovachki2023neural,kovachki2021universal,kovachkiChapter9Operator2024}. Moreover, once NOs are trained,  they can be evaluated highly efficiently in a single forward pass. However, despite these features, purely data-driven NOs still suffer from the data challenge: they require a large set of paired data and do not generalize well when test samples are out of the training regime. To reduce this generalization gap, PINO \citep{liPhysicsInformedNeuralOperator2024a} and physics-informed DeepONet \citep{goswami2023physics,wang2021learning} were introduced, where a PDE-based loss is added to the training loss as a penalization term. When applying the PDE loss in the test phase, a test-time optimization is performed to take advantage of both the learned neural operator and the additional governing equation, enabling a more accurate solution function and a smaller generalization error on the querying instance \citep{liPhysicsInformedNeuralOperator2024a}. However, computing derivatives of batched outputs through backpropagation demands a large memory overhead. Moreover, the highly complex landscape of the physics-informed loss leads to a long optimization procedure, even the failure of convergence. And one needs to carefully fine-tune the loss weights of the data loss and the PDE loss to enhance the performance. 

\subsection{Physics-Informed Neural Network}\label{subsec:bg-pinn}

PINNs have become a powerful tool for solving various PDEs arising from numerous forward and inverse problems~\cite{raissi2019physics}. Instead of approximating the operator between function spaces, PINNs employ deep neural networks trained on several sets of collocation points in the physical domain and its boudaries to approximate the solutions to a given PDE by minimizing the loss that consists of the PDE residual as well as boundary/initial conditions. Specifically, assume that the solution $u\in\mathcal U$ satisfies the following time-dependent PDE defined on a bounded domain $\Omega\subset\mathbb{R}^m$ and its boundary $\partial\Omega$
    \begin{align}
            \label{eq:pinn-pde}
        \pdv{u}{t}+(\mathcal{N}u)(\bm{x},t)&=g(\bm x,t), && (\bm x,t) \in\Omega\times[0,T],\\
        (\mathcal{B}u)(\bm x,t)&=0, && (\bm x,t)\in\partial\Omega\times[0,T],\\
        u(\bm x,0)&=u_0(\bm x), && \bm x\in\Omega,
    \end{align}
    where $\mathcal{N}$ encodes a differential operator and $\mathcal{B}$ the boundary conditions. 
    The PINN solution $u_{\rm NN}(\bm x,t;\theta)$ parameterized by neural network parameters $\theta\in\Theta$ is identified by minimizing the loss function: 
    \begin{align}
        \mathcal{L}[u_{\rm NN}(\bm x,t;\theta)] = &\left\|\pdv{u_{\rm NN}}{t}+(\mathcal{N}u_{\rm NN})-g\right\|_{L^2}^2 + \|(\mathcal{B}u_{\rm NN})\|_{L^2}^2 + \|u_{\rm NN}-u_0\|_{L^2}^2 \label{eq:pinn-loss}
    \end{align}
    The $L^2$ norms for the three terms above are defined in the domains $\Omega \times [0,T]$, $\partial\Omega \times [0,T]$, and $\Omega \times \{t=0\}$, respectively. This loss can be discretized and evaluated on three sets of collocation points $\{(\bm{x}_i^R,t_i^R\}_{i=1}^{s_R}$ for the residual, $\{(\bm{x}_i^B, t_i^B\}_{i=1}^{s_B}$ for the boundary condition, and $\{\bm{x}_i^I\}_{i=1}^{s_I}$ for the initial condition.

However, the loss \cref{eq:pinn-loss} takes the strong form of the PDE causing PINN to possibly fail to provide accurate solutions to the PDEs with weak regularity. Therefore, we resort to the so-called Variational PINN (VPINN) first proposed in~\cite{kharazmiVariationalPhysicsInformedNeural2019a}, which applies the weak form of the PDE when calculating the residual loss. Let the weak form of the PDE be $a(u,v)=\ell(v)\ (v\in \mathcal V)$ where $\mathcal V$ denotes the test function space and $\ell(v)$ corresponds to the source term $g$ in \cref{eq:pinn-pde}. In this work, we use the loss from the formulation of Robust VPINN~\cite{rojasRobustVariationalPhysicsInformed2024}
    \begin{equation}\label{eq:rvpinn-loss}
        \mathcal{L}[u_{\rm NN}(\bm{x},t;\theta)]=R(u_{\rm NN})^TG^{-1}R(u_{\rm NN}) + \mathcal{L}_b(u_{\rm NN}) + \mathcal{L}_i(u_{\rm NN})
    \end{equation}
where $R$ is a vector-valued function, the element of which is defined as the residual of the weak form with respect to a set of test functions $\{\varphi_i(x)\in \mathcal V\}_{i=1}^{N_f}$. That is,  $R(u)_n\coloneqq r(u,\varphi_n)\coloneqq \ell(\varphi_n) - a(u,\varphi_n) \ (n=1,2,\dots, N_f)$. $G$ is the Gram matrix and is defined by $G_{mn}\coloneqq (\varphi_n,\varphi_m)_{\mathcal V}$ ($(\cdot, \cdot)_{\mathcal V}$ denotes the inner product in the function space $\mathcal V$ and $\|\cdot\|_{\mathcal V}$ the norm induced by the inner product). $\mathcal{L}_b$ and $\mathcal{L}_i$ are the same losses in \cref{eq:pinn-loss} of the boundary conditions and the initial conditions. \par 

Although PINN and its variants have obtained promising results on various physics modeling tasks, PINNs face several challenges. (1) For problems involving complex physics \citep{fuks2020limitations,raissi2020hidden,wang2022and}, the PDE constraint induces highly nonlinear optimization landscapes \citep{wang2021understanding,gopakumar2023loss}, resulting in slow or even failed convergence.
(2) For every new instance of boundary conditions/physical parameters, a new training needs to be performed in PINNs. As a result, PINNs become expensive when multiple instances of PDEs need to be solved, such as in inverse PDE and design problems \citep{lee2024data,smith2022hyposvi}. To alleviate the second challenge, several strategies have been proposed. Examples include the TL-PINN approach based on a transfer-learning philosophy to fine-tune a pre-trained model and fit the new physics constraint \citep{xu2023transfer,prantikos2023physics,desai2021one,goswami2020transfer}, the GPT-PINN approach which leverages a meta-learning idea to adaptively infer the parametric dependence of the system and expands a meta-network \citep{chen2024gpt,chenTGPTPINNNonlinearModel2024a,koumpanakis2024meta,zhangMetaNOHowTransfer2023}, and so on. However, to the best of our knowledge, none of these approaches is capable of resolving the optimization challenge in PINNs.

\subsection{Reduced Basis Method}\label{subsec:bg-rbm}

As ReBaNO is inspired by RBM, we devote this subsection to a brief review of RBM. RBM \cite{HesthavenRozzaStammBook, quarteroni_reduced_2016, Haasdonk2017Review} is a popular model order reduction (MOR) technique capable of rigorously and efficiently simulating parametric PDEs. Its signature, in comparison to other MOR approaches, is a greedy algorithm integrated in an offline-online learning procedure. The offline (i.e., training) stage is devoted to a judicious exploration of the parameter-induced solution manifold  driven by an error estimator. It relies on a high-fidelity numerical procedure for the parameter to solution map $\mu \rightarrow u(x; \mu) \in X_h$. We call this Full-Order Model, FOM$({\mu}, X_h)$. RBM selects a number of representative parameter values $\{\mu^n\}_{n=1}^N$ via a mathematically rigorous greedy algorithm \cite{BinevCohenDahmenDevorePetrovaWojtaszczyk} to form an $N$-dimensional subspace of $X_h$, $X_N \coloneqq \mathrm{span}\left\{u(\cdot; \mu^n)\right\}_{n=1}^N$. During the online stage, the method computes an approximation $u_N(\cdot, \mu)$ in the surrogate space for any unseen (but still in distribution) parameter value via the ansatz $u_N(\cdot, \mu) = \sum_{n=1}^N c_n(\mu) u(\cdot, \mu^n)$.  Thanks to this ansatz and the greedy algorithm iteratively called to build the surrogate space from ground up, in comparison to other reduction techniques (e.g. proper orthogonal decomposition (POD)-based approaches \cite{Kunisch_Volkwein_POD, BennerGugercinWillcox2015}), the number of full order inquiries RBM takes offline is minimum, i.e., equal to the dimension of the surrogate space.

\section{A Unified Operator Learning Framework}\label{sec:unified-ol-framework}
In this section, we aim to outline a unified operator learning framework to place our discussions in context. In \cref{subsec:olfw-problem-setting}, we first introduce the notation and establish the problem setting. In \cref{subsec:olfw-olfw}, we elucidate the details of the framework and give a unified description of the NOs that we study in our experiments. 

\subsection{Problem Setting}\label{subsec:olfw-problem-setting}
Without losing generality, we consider a family of PDEs of the following general form:
\begin{align}
    \pdv{u}{t} + (\mathcal{N}_fu)(\bm x,t) &= g(\bm x, t), && (\bm x,t)\in\Omega\times[0,T] \label{eq:pde}\\
    (\mathcal{B}u)(\bm x,t) &= h(\bm x,t), && (\bm x,t)\in\partial \Omega \times [0,T] \label{eq:pde-bc}\\
    u(\bm x,0) &= u_0(\bm x), && \bm x\in\Omega \label{eq:pde-ic}
\end{align}
for some $h\in\mathcal{U}_b$, $u_0\in\mathcal{U}_i$ and $\Omega\subset\mathbb{R}^n$ (for some positive integer $n$) a bounded domain, where $\mathcal{N}_f:\mathcal U\times \mathcal F\to \mathcal U^*$ is a (non)linear differential operator dependent on an input function $f\in\mathcal{F}$ and $\mathcal{B}$ encodes the boundary condition. Assume that the solution $u:\Omega\times[0,T]\to\mathbb{R}^{d_u}$ belongs to a Banach function space $\mathcal{U}$, and that the boundary function space $\mathcal{U}_b$ and the initial function space $\mathcal U_i$ are also Banach spaces. The source term $g$ on the right-hand side is from the dual space of $\mathcal{U}$, denoted by $\mathcal{U}^*$. We also assume that the input function space $\mathcal{F}$ is a Banach space of functions and is defined as
$\mathcal{F}\coloneq\{f:\Omega'\to\mathbb{R}^{d_f},\ \Omega'\subset\mathbb{R}^{m}\}$ (for some positive integer $m$).\footnote{We remark that, for notational simplicity, we focus on variations with respect to $f$ only, though the formulation naturally extends to learning operators involving variations of $g$ and $h$ as well.} For simplicity of our discussion, we assume $\Omega'=\Omega$. That means, the input function (e.g., parameter field) and the solution are defined on the same physical domain. \par 
The goal of operator learning is to approximate the groundtruth solution operator $\Psi$ mapping between two infinite-dimensional Banach spaces $\mathcal{F}$ and $\mathcal{U}$, i.e., $\Psi:\mathcal{F}\to\mathcal{U}$. To this end, we construct a parametric operator
\begin{equation}
    \Psi^\sim:\mathcal{F}\times\Theta\to\mathcal{U}
\end{equation}
where $\Theta$ is the trainable parameter space and we seek an optimal $\theta^*\in\Theta$ so that the resulting operator $\Psi^*(\cdot)\coloneq\Psi^\sim(\cdot,\theta^*)\approx\Psi(\cdot)$. The optimal parameter $\theta^*$ is found by minimizing the loss function $\mathcal{L}(\Psi,\Psi^\sim;\theta)$, i.e.
\begin{equation}
    \theta^* = \mbox{\rm arg}\hspace*{-1pt}\min_{\theta \in\Theta} \mathcal{L}(\Psi,\Psi^\sim;\theta).
\end{equation}
For data-driven models, we define the loss function as the Bochner norm of the approximation error
\begin{equation}\label{eq:no-loss}
    \mathcal{L}_d(\Psi,\Psi^*;\theta)\coloneqq\mathbb{E}_{f\sim\mu}\,\|\Psi(f)-\Psi^\sim(f;\theta)\|_{L^2_\mu(\mathcal{F};\mathcal{U})}^2=\int_{\mathcal{F}}\|\Psi(f)-\Psi^\sim(f;\theta)\|_{\mathcal{U}}^2\dd{\mu(f)},
\end{equation}
where $\mu$ is a probability measure supported on $\mathcal{F}$.
To compute this loss, we need $N_p$ i.i.d. Monte Carlo samples of the input functions $\{f_i\}_{i=1}^{N_p}$ from $\mu$ and possibly a set of corresponding (approximate) solutions $\{u_i\}_{i=1}^{N_p}$ with $u_i \approx \Psi(f_i)$ for training.
Thus, the loss can be approximated by the empirical risk.
\begin{equation}\label{eq:no-emp-risk}
    \mathcal{L}_d(\Psi,\Psi^*;\theta)\approx\frac{1}{N_p}\sum_{i=1}^{N_p}\|u_i-\Psi^\sim(f_i;\theta)\|^2_{\mathcal{U}}.
\end{equation}
This empirical risk is usually computed using the $L^2$ relative error instead of the $\mathcal{U}$-norm. \par 
For physics-driven models, the loss function is independent of the approximated solution data and consists of the PDE residual along with the boundary and the initial condition
\begin{align}\label{eq:physics-loss}
    \mathcal{L}_p(\Psi^\sim;\theta) \coloneqq \mathbb{E}_{f\sim\mu}\,&\bigg[\bigg\|\pdv{\Psi^\sim(f;\theta)}{t}+\mathcal{N}_f\Psi^\sim(f;\theta) - g\bigg\|_{\mathcal{U}^*}  + \|\mathcal{B}\Psi^\sim(f;\theta)-h\|_{\mathcal{U}_b} + \|\Psi^\sim(f;\theta)-u_0\|_{\mathcal{U}_i}\bigg].
\end{align}
The above three norms are defined over the domains $\Omega\times[0,T]$, $\partial\Omega\times[0,T]$ and $\Omega\times\{t=0\}$, respectively. This loss function is exactly the same as the PINN loss \cref{eq:pinn-loss} when the norms are replaced with the $L^2$ norm. Although there has been work indicating that proper norms should be chosen for convergence and error control (see, for example,~\cite{bonitoConvergenceErrorControl2025}), we adopt the $L^2$ norm in our experiments for simplicity and practical use. In this case, the empirical risk of the physics-informed loss becomes 
\begin{align}\label{eq:physics-l2-loss}
    \mathcal{L}_p(\Psi^\sim;\theta) \approx \frac{1}{N_p}\sum_{i=1}^{N_p}&\bigg[\bigg\|\pdv{\Psi^\sim(f_i;\theta)}{t}+\mathcal{N}_f\Psi^\sim(f_i;\theta) - g\bigg\|_{L^2}^2 \nonumber \\
    &\quad + \|\mathcal{B}\Psi^\sim(f_i;\theta)-h\|_{L^2}^2 + \|\Psi^\sim(f_i;\theta)-u_0\|_{L^2}^2\bigg].
\end{align}
\par 
In practice, the function data of $\{f_i\}_{i=1}^{N_p}$, $\{u_i\}_{i=1}^{N_p}$ and $\{g,h,u_0\}$ are always measured or evaluated at specific locations in the spatial or spatio-temporal domain. To be concrete, we discretize the domains into three sets of collocations points of $\Omega\times[0,T]$, $\partial\Omega\times[0,T]$ and $\Omega\times\{t=0\}$ of size $s_R$, $s_B$ and $s_I$, respectively. Therefore, we define
\begin{align}
    \mathcal{C}_R&=\{(\bm x_i,t_i)\in\Omega\times[0,T]\}_{i=1}^{s_R}, \\
    \mathcal{C}_B&=\{(\bm x_i,t_i)\in\partial\Omega\times[0,T]\}_{i=1}^{s_B}, \\
    \mathcal{C}_I&=\{\bm x_i\in\Omega\}_{i=1}^{s_I}.
\end{align}
In this setting, we assume that the data for $\{f_i, u_i\}_{i=1}^{N_p}$ and $g$ are available on $\mathcal{C}_R$ (although $f_i$ and $u_i$ are not necessary to be evaluated at the same locations), for $h$ on $\mathcal{C}_B$ and for $u_0$ on $\mathcal{C}_I$ so that the loss \cref{eq:no-emp-risk} and \cref{eq:physics-l2-loss} can be evaluated at the corresponding sites. \par 
With this problem setup, we attempt to give a unified description of operator learning models.

\subsection{Encoder-Decoder Framework}\label{subsec:olfw-olfw}
\begin{figure}[!htbp]
\centering
\begin{tikzcd}[row sep=large, column sep=normal]
  & \mathcal U_N \arrow[dr, "\mathcal{D}_o"] & \\
  \mathcal F \arrow[ur, "\mathcal{E}_i"] \arrow[rr, "\Psi"'] 
  & & \mathcal U
\end{tikzcd}
\caption{The key workflow of operator learning models. $\Psi$ denotes the solution operator to be approximated. The encoder $\mathcal{E}_i$ maps the input space $\mathcal{F}$ to the latent space $\mathcal{U}_N$, and the decoder $\mathcal{D}_o$ maps $\mathcal{U}_N$ back to the output space $\mathcal{U}$.}
\label{fig:ol-model-workflow}
\end{figure}
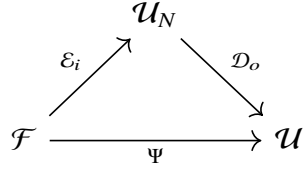

In this subsection, we unify all relevant operator learning models within a common framework. 
Many operator learning models adopt an autoencoder-like structure, where the central idea is to approximate the infinite-dimensional output $u \in \mathcal{U}$ through a latent finite-dimensional representation $u_N \in \mathcal{U}_N$, where $\mathcal{U}_N$ is a learned $N$-dimensional latent space. Accordingly, the key task is to capture this latent structure of the output space and approximate the mapping from the input function $f \in \mathcal{F}$ to $u_N \in \mathcal{U}_N$.
Therefore, the architectures of many operator learning models can be split into two components: (1) an encoder $\mathcal{E}_i:\mathcal{F} \to \mathcal{U}_N$ that maps the input function to its latent representation, and
(2) a decoder $\mathcal{D}_o:\mathcal{U}_N \to \mathcal{U}$ that lifts the latent representation back to the full output space. 
That is, the target operator $\Psi:\ \mathcal{F}\to\mathcal{U}$ can be approximated as
\begin{equation}\label{eq:olfw}
    \Psi\approx \mathcal{D}_o\circ \mathcal{E}_i,
\end{equation}
see \cref{fig:ol-model-workflow} for a schematic plot. This encoder-decoder decomposition is the abstraction of the architectures of many neural operators. It extends the applications of autoencoders to learning mappings between infinite-dimensional function spaces.

Note that operator learning models can only learn the corresponding discrete representations of these operators given discretizations $\mathcal{C}_R$, $\mathcal{C}_B$ and $\mathcal{C}_I$. Ideally, the learned discrete operators should be alias-free and preserve \textit{continuous-discrete-equivalence} for robust operator learning, meaning that the continuous solution can be recovered from the discrete output and the discrete output maintains the continuous properties. But we are not going to address this issue in this work, we refer to~\cite{bartolucciRepresentationEquivalentNeural2023a} for more detailed discussion about alias-free operator learning. In the following discussions, we also use these notations for discrete representations of the input encoder and the output decoder. \par

In this work, we consider four representative operator models for comparison: PCA-Net, DeepONet, FNO and CNO. In the following, we will present the descriptions of how they fit into the Encoder-Decoder framework \cref{eq:olfw}. For an overview, we refer the reader to \cref{tab:olfw}.\par 

\begin{table}[htbp]
    \centering
    \begin{threeparttable}
        \caption{An overview of the selected models under the Encoder-Decoder framework}
        \label{tab:olfw}
        \begin{tabular}{ccc} 
            \toprule
            Model & Encoder ($\mathcal{E}_i$) & Decoder ($\mathcal{D}_o$) \\
            \midrule
            PCA-Net & PCA + NN        & linear             \\
            DeepONet & NN        & linear            \\
            FNO      & linear layer + Fourier layers & linear layer \\
            CNO      & linear layer + composition of $D,R,I,U$ blocks & linear layer \\
            ReBaNO   & A single-layer NN loss minimizer & \makecell[c]{linear decoder \\ linear, via greedy selected PINNs} \\
            \bottomrule
        \end{tabular}
        \begin{tablenotes}[flushleft]
            \item[*] \small PCA: principal component analysis; NN: neural network
        \end{tablenotes}
    \end{threeparttable}
\end{table}

\paragraph{PCA-Net} To deliver this paragraph properly, we assume that $\mathcal{F}$ and $\mathcal{U}$ are also generic real separable Hilbert spaces equipped with inner product $(\cdot,\cdot)_{\mathcal{F}}$ and $(\cdot,\cdot)_{\mathcal{U}}$. PCA-Net applies Principal Component Analysis (PCA) onto the data pairs $\{f_i, u_i\}_{i=1}^{N_p}$ evaluated on $\mathcal{C}_R$, generating a reduced basis $\{\phi_i\}_{i=1}^M$ for the input space $\mathcal F$ and a reduced basis $\{\psi_i\}_{i=1}^N$ for the output space $\mathcal U$ for some predefined positive integers $M$ and $N$. Then all $\{f_i, u_i\}_{i=1}^{N_p}$ are projected onto these bases through
\begin{align}
    \alpha_{ki}&\coloneqq (\phi_k, f_i)_{\mathcal F},\ k=1,2,\dots,M \\
    \beta_{li}&\coloneqq(\psi_l,u_i)_{\mathcal U}, l=1,2,\dots,N
\end{align}
for $i=1,2,\dots,N_p$. Then PCA-Net uses a feedforward neural network $\Psi_{\rm NN}^{\rm PCA}:\mathbb R^M \times\Theta\to \mathbb R^N$ to learn the mapping between the PCA coefficients by feeding the data $\{\boldsymbol\alpha_i\in\mathbb R^M\}_{i=1}^{N_p}$ and $\{\boldsymbol\beta_i\in\mathbb R^N\}_{i=1}^{N_p}$. Let the operator $\mathcal R:\mathcal F\to\mathbb R^M$ and $\mathcal S:\mathbb R^N\to\mathcal U$ be defined as
\begin{align}
    \mathcal R(f)&\coloneqq\boldsymbol\alpha(f)=\{\langle\phi_k, f\rangle_{\mathcal F}\}_{k=1}^M \label{eq:pca-proj}\\
    \mathcal S(\boldsymbol\beta)&\coloneqq\sum_{l=1}^N\beta_l\psi_l\label{eq:pca-decoder}
\end{align}
for any $f\in\mathcal F$ and $\boldsymbol\beta\in\mathbb R^N$. Then, for a given $f\in\mathcal F$, the prediction given by PCA-Net is
\begin{equation}
    \Psi^\sim_{\rm PCA}(f;\theta) = \sum_{l=1}^N\Psi_{\mathrm{NN}, l}^{\rm PCA}(\mathcal Rf;\theta)\psi_l,
\end{equation}
where the subscript $l$ in $\Psi_{\mathrm{NN}, l}^{\rm PCA}$ denotes the $l$-th component.
Therefore, the input encoder of PCA-Net is 
\begin{equation}\label{eq:pca-encoder}
    \mathcal E_i^{\rm PCA}=\Psi_{\rm NN}^{\rm PCA} \circ \mathcal R
\end{equation}
while the output decoder $\mathcal D_o$ is exactly the operator $\mathcal S$ defined in \cref{eq:pca-decoder}.

\paragraph{DeepONet} DeepONets consist of two trainable neural networks, a branch net and a trunk net. The branch net encodes the input function, while the trunk net outputs temporally dependent basis functions that span the solution space. In the original DeepONet~\cite{lu2021learning}, the branch net $\Psi_{\rm NN}^b$ takes as input the discrete evaluations of the input function at fixed locations, $\tilde{f} \coloneq [f(x_1), f(x_2), \dots, f(x_M)] \in \mathbb{R}^M$, and maps them to an $N$-dimensional latent vector $\mathbf b(\tilde f)=\{b_i(\tilde f)\}_{i=1}^N$ via
\begin{equation}\label{eq:deeponet-encoder}
    \Psi_{\rm NN}^b:\mathbb R^M\times\Theta\to\mathbb R^N.
\end{equation}
In our implementation, we follow an improved version of the branch net~\cite{hoopCostAccuracyTradeOffOperator2022a,lu2022comprehensive}, where a PCA transform is first applied to the input function $f$, and the branch net learns the mapping from the PCA coefficients to the latent representation:
$$
\mathbf b=\Psi_{\mathrm{NN}}^{\rm PCA}(\mathcal Rf;\theta),
$$ 
which is identical to the PCA-Net.
The trunk net $\Psi_{\rm NN}^{t}:\Omega\times\Theta\to\mathbb R^N$ outputs the basis functions for reconstructing the solution. 
If we let $\boldsymbol \psi_{\rm NN}(\cdot;\theta^t)\coloneq\Psi_{\rm NN}^t(\cdot;\theta^t):\Omega\times\Theta\to\mathbb R^N$, combining the output of the branch net, the prediction given by DeepONet is
\begin{equation}
    \Psi^\sim_{\rm DON}(f;\theta)(\bm x,t) = \sum_{l=1}^N\Psi_{\mathrm{NN}, l}^{\rm PCA}(\mathcal R f;\theta^b)\psi_{\mathrm{NN}, l}(\bm x,t;\theta^t),\ (\bm x,t)\in\Omega\times[0,T]
    \label{eq:deeponet}
\end{equation}
where $\theta$ is the collection of $\theta^b$ and $ \theta^t$. This architecture shares the same encoder-decoder structure as PCA-Net.

\paragraph{FNO} Contrary to PCA-Net and DeepONet, FNO is a kernel-based approach that leverages integral kernel layers to capture local and nonlocal features. By definition (see Eq. (6) in~\cite{kovachki2023neural}), the architecture can be divided into three components.
\begin{equation}\label{eq:fno-arch}
    \Psi^\sim_{\rm FNO}(f;\theta)=\mathcal{Q}\circ\mathcal K\circ\mathcal{P}(f)
\end{equation}
where the lifting layer $\mathcal{P}$ is a learnable linear layer, defined in a pointwise fashion as $\mathcal{P}(f)\coloneq W_0f+b_0$ for some $W_0\in\mathbb R^{d_f'\times d_f}$ and $b_0\in\mathbb R^{d_f'}$. It ``lifts" the input function space $\mathcal{F}=\{f:\Omega\to\mathbb R^{d_f}\}$ to a higher-dimensional latent feature space $\mathcal{F}_r\coloneq\{f:\Omega\to\mathbb R^{d_f'}\}$ for $d_f'>d_f$. Then the input will go through several consecutive Fourier layers
\begin{equation}
    \mathcal K=\mathcal{K}_{L}\circ\mathcal{K}_{L-1}\circ\cdots\circ\mathcal{K}_2\circ\mathcal{K}_1
\end{equation}
for some positive integer $L$. For $0<l<L$, a single layer $\mathcal{K}_l:v_l\mapsto v_{l+1}$is pointwise defined as
\begin{equation}
    v_{l+1}(y)=\sigma_{l+1}\pqty(W_{l+1}v_l(y)+\int_\Omega\kappa_l(x-y)v_l(x)\dd{x} + b_l), \ y\in\Omega
\end{equation}
where $\sigma$ is a nonlinear activation function and $\{W_l,b_l,\kappa_l\}_{l=1}^L$ are weights, biases, and parametric kernels. The computation of the convolution integral is carried out in the Fourier domain. Then, the projection layer $\mathcal{Q}$ projects the output $v_L$ back into the solution space $\mathcal U$, defined by $u_{\rm FNO}=W_{L+1}v_L+b_{L+1}$. Therefore, the encoder of FNO is the composition of the lift layer and a set of Fourier layers
\begin{equation}
    \mathcal E_i^{\rm FNO}=\mathcal{K}\circ \mathcal P
\end{equation}
while the decoder is the projection layer, i.e.,
\begin{equation}
    \mathcal D_o^{\rm FNO} = \mathcal Q
\end{equation}
As such, the parameter $\theta$ of FNO is the collection of all learnable neural network parameters, weights and biases $\{W_l,b_l\}_{l=0}^{L+1}$ and parametric kernels $\{\kappa_l\}_{l=1}^L$. \par

\paragraph{CNO} CNO is also a kernel-based neural operator. It can be viewed as a modified U-Net~\cite{ronnebergerUNetConvolutionalNetworks2015}. Similar to FNO, it also includes a lifting layer $\mathcal{P}$ and a projection layer $\mathcal{Q}$. However, the layers inside learn the input encoder $\mathcal E_i$ in a very different way. For a given input, it will successively go through downsampling blocks ($D$ blocks), residual blocks ($R$ blocks), invariant blocks ($I$ blocks) and upsampling blocks ($U$ blocks). Such architecture effectively helps the neural network to capture the local features by downsampling and convolutional operations in the downsampling blocks and reconstruct the solution by upsampling and concatenating with the global features through the residual connections (see \cite{raonicConvolutionalNeuralOperators2023a} for more details). Hence, the encoder of CNO is the composition of a lift layer and a set of $D,R,I,U$ blocks and the projection layer works as the decoder. \par

It is worth mentioning that the input pairs are evaluated on a discrete grid of size $s_R$ points. Therefore, the PCA input encoder \cref{eq:pca-encoder} depends on the mesh size $s_R$, and therefore PCA-Net and DeepONet are not resolution invariant on the input end. In opposite, each layer of FNO works as an operator and the operations involved are resolution independent, and for inputs with different resolutions, CNO down/upsamples both input and output data to a fixed resolution, so CNO is also invariant with respect to the input/output resolutions. \par 
After the problem setup and framework building, we are ready to introduce ReBaNO in the next section. 

\section{Reduced Basis Neural Operator (ReBaNO)}\label{sec:rebano}
\label{sec:methodology}
We devote this section to a systematic description of the methodology for the proposed neural operator. In contrast to traditional data-driven operator learning models, ReBaNO relies on less high-fidelity data thanks to its judicious greedy strategy in determining which specific high-fidelity data to leverage. 
Moreover, ReBaNO fine-tunes the model online when a new input function is presented by injecting the corresponding physics as opposed to simply evaluating the trained map. It is these two features that lead to ReBaNO possessing stronger generalizability and mesh invariance.

\subsection{The Key Idea of ReBaNO}
\begin{figure*}[htbp]
    \centering
    \includegraphics[width=0.9\linewidth]{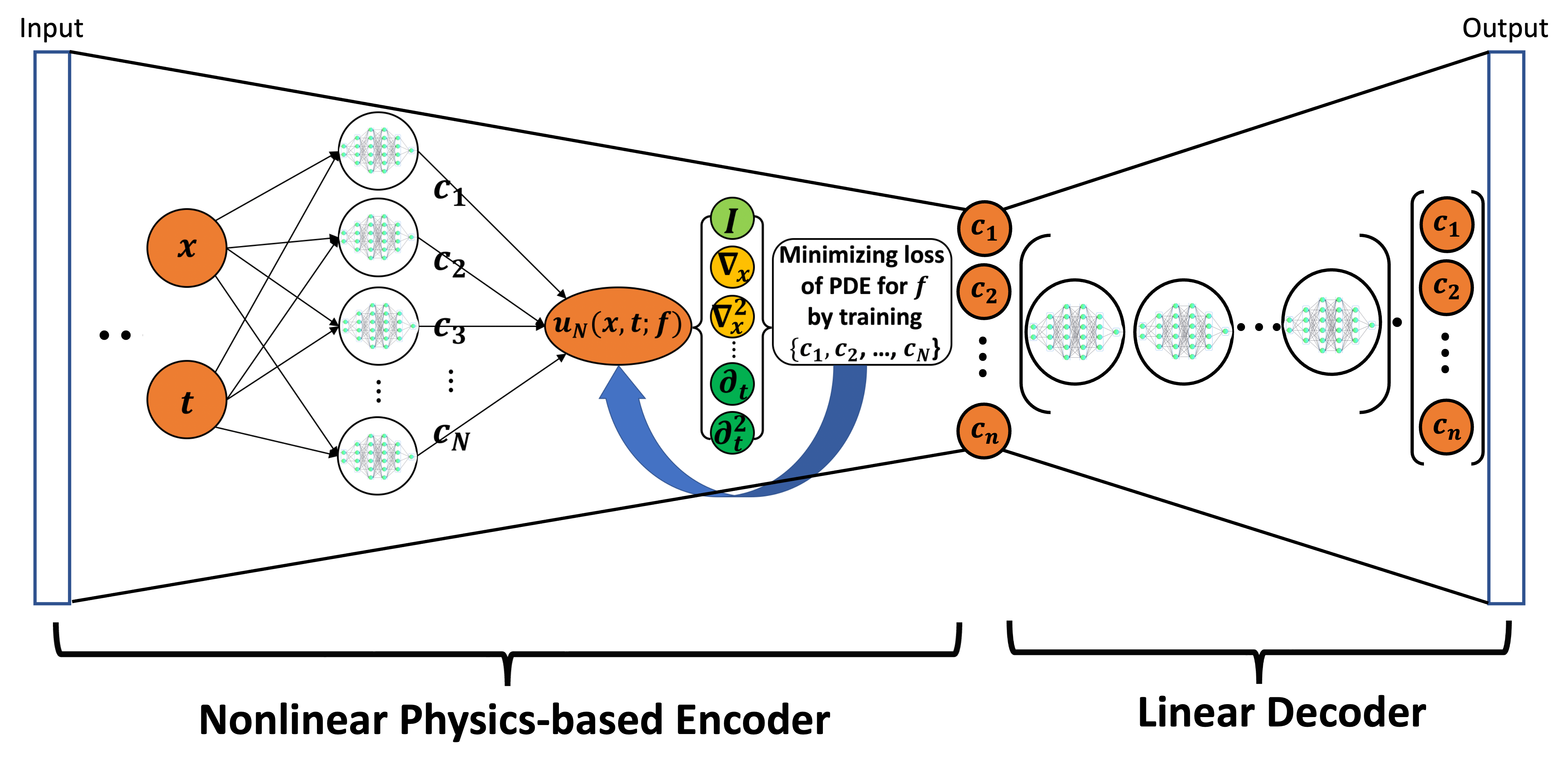}
    \caption{Schematics of ReBaNO. Given an input function, the nonlinear encoder encodes it to $\bc$ which then is decoded to the output by a linear operator. Both encoder and decoder are adaptively generated from scratch through an offline training process allowing this minimalist design.}
    \label{fig:rebano}
\end{figure*}

Inspired by RBM and GPT-PINN, the Reduced Basis Neural Operator adopts a nonlinear physics-based encoder followed by a linear decoder (\cref{fig:rebano}). This nonlinear-linear pattern was proven to be more theoretically advantageous than other patterns \cite{BUCHFINK2024134299}. It builds the mapping in an offline-online fashion, relying on high-fidelity PINN solutions trained on a small number of data instances. In fact, to learn the operator
$\Psi: \mathcal{F} \rightarrow \mathcal{U}$, ReBaNO iteratively identifies $N$ input-output pairs via a greedy algorithm $\{(f_i(\bm x,t), u_i(\bm x,t) = \Psi_h(f_i)(\bm x,t))\}$ offline. Here, $\Psi_h$ is a high-fidelity approximation of $\Psi$ capable of resolving the input-output map at least at certain input points. ReBaNO builds a rank-$n$ ($1 \le n \le N$) approximation of $\Psi$ in the following fashion: for an arbitrary input function $f\in\mathcal{F}$, 
this rank-$n$ ReBaNO parameterize the operator $\Psi_n: \mathcal{F} \times \Theta \rightarrow \mathcal{U}$ by $\bc \eqqcolon \{c_i\}_{i=1}^n \in \mathbb R^n$
\begin{equation}\label{eq:rebano-ansatz}
   \Psi_n(f)(\bm x,t)= \tilde{u}(\bm x,t; \bc)\coloneqq \sum_{i=1}^n c_iu_i(\bm x,t).
 \end{equation}
In essence, ReBaNO adopts a simple one-layer instance-wise network with zero bias, customized activation functions $ \{u_i\}_{i=1}^n$, and $\bc \coloneqq \{c_i\}_{i=1}^n$ being the trainable weights.

\subsection{Physics-Driven Online Tuning}
These weights $\{c_i\}_{i=1}^n$ are tuned online by ensuring that the physical law corresponding to the map $f(x,t) \mapsto u(x,t) = \Psi(f)(x,t)$ is satisfied. The ReBaNO solution is sought to satisfy the same PDE \cref{eq:pde} and boundary/initial conditions. The ReBaNO parameter $\bc(f)$ is identified per instance by minimizing the loss function \cref{eq:physics-l2-loss} for $N_p=1$. 

We remark that, thanks to the design of offline-online framework and that the activation functions $\{u_i(x,t)\}_{i=1}^n$ are pre-trained and the $\bc$-dependence is linear, the relevant derivatives in the loss function can be computed and stored in advance. The consequence is that the weights $\{c_i\}_{i=1}^n$ can be obtained rapidly. In fact, when $\mathcal{N}_f$ in \cref{eq:pde} is a linear operator, no optimization method is needed, as it simplifies to a small least squares problem.

\subsection{Knowledge Distillation via Customized Activations}

\begin{algorithm}[h]
    \caption{ReBaNO: Greedy Knowledge Distillation}
{\bf Input: }{Discrete set $F$ for input $f$, high-fidelity approximate operator $\Psi_h$ (PINN solver), training loss $\mathcal{L}_p$}
\begin{algorithmic}[1]
\STATE Use the high-fidelity solver to obtain $u_1$ at a random input $f_1$, $u_1 = \Psi_h(f_1)$. Precompute quantities necessary for quickly evaluating \cref{eq:physics-l2-loss}. Set $n=2$.
\WHILE{{\em stopping criteria not met,}}
\STATE Train the rank-$(n-1)$ ReBaNO $\Psi_{n-1}(f)$ (i.e., solving the corresponding latent code $\bc \in \mathbb{R}^{n-1}$) for all input $f$ in the discretized input set and record the indicator 
$\Delta_{n-1}(f) \coloneqq \min_{\bc}\mathcal{L}_p[\Psi_{n-1}(f);\bc]$.
 \STATE Choose $f_n = \displaystyle
  \mbox{\rm arg}\hspace*{-1pt}\max_{f \in{F}}
 {\Delta_{n-1}(f)}$.
\STATE  Use the high-fidelity solver to obtain $u_n$ at $f_n$, $u_n = \Psi_h(f_n)$. Precompute quantities necessary for quickly evaluating \cref{eq:physics-l2-loss}.
\STATE Update the ReBaNO network by adding a neuron to the hidden layer to construct $\Psi_n(f)$, and set $n \leftarrow n+1$.
\ENDWHILE
\end{algorithmic} 
{\bf Output:} Sequence of operators $\{\Psi_n: 1 \le n \le N\}$ approximating the unknown operator $\Psi$.
    \label{alg:rebanogreedy}
\end{algorithm}
ReBaNO features a network-of-networks design with the outer network distilling knowledge from the inner ones. During the offline stage, the outer network grows one neuron at a time self-adaptively by selecting a representative input via a Greedy algorithm. 
Given a set of samples $F=\{f_i\}_{i=1}^{N_p}$ through which ReBaNO investigates during the training phase, let the set of selected input functions be $\{f_i:f_i\in F\}_{i=1}^n$ and their corresponding output (PINN solutions to the PDE) be $\{u_i^\mathrm{PINN}\}_{i=1}^n$. 
The ReBaNO network \cref{eq:rebano-ansatz} amounts to adopting these problem-dependent networks $\{u_i^\mathrm{PINN}\}_{i=1}^n$ as its activation functions. Such an adaptive architecture and the knowledge distillation enable automatic learning of the latent low-dimensional structure of the to-be-learned map. 

We end this subsection by describing the iterative and greedy procedure leading to these customized activations. Its main steps are outlined in Algorithm \ref{alg:rebanogreedy}.  
The ReBaNO network adaptively ``learns'' the input-output relations and ``grows'' its sole hidden layer one neuron/network at a time in the following fashion. We first randomly select one input instance and obtain the associated (highly accurate) output $u_1(x,t)$. The algorithm then decides how to ``grow'' itself by scanning the entire discrete input space and, for each case, training this reduced operator $\Psi_1$ (of 1 hidden layer with 1 neuron). As it scans, it records an error indicator - terminal loss of $\Psi_1(f)$. The next chosen input instance is the one generating the largest error indicator. The algorithm then proceeds by obtaining the high-fidelity output for that input using a PINN solver and therefore grows $\Psi_1$ to be $\Psi_2$ featuring two neurons with customized (but pre-trained) activation functions. At every step, we select the input instance that is approximated most badly by the current approximate operator $\Psi_n$. 

\subsection{ReBaNO Under the Encoder-Decoder Framework}
After introducing the building blocks of ReBaNO, we end this section with the Encoder-Decoder decomposition of ReBaNO. ReBaNO constructs a surrogate output space $\mathcal{U}_N$ spanned by the reduced basis $\{u_i^{\rm PINN}(\cdot;\theta_i)\}_{i=1}^N$ (here every $\theta_i\in\Theta_{\rm PINN}$ denotes the set of network parameters of each PINN) by distilling knowledge of the full PINN solutions by a greedy algorithm during offline training. Hence, in the terminology of \cref{sec:unified-ol-framework}, the ReBaNO output decoder $\mathcal{D}_o^{\rm ReBaNO}$ is learned by physics-driven greedy selection. The output decoder is linear and explicitly defined as the linear combination of the reduced basis
\begin{equation}\label{eq:rebano-decoder}
    \mathcal{D}^{\rm ReBaNO}_o(\bc)(\bm x,t)\coloneq\sum_{i=1}^Nc_iu_i^{\rm PINN}(\bm x,t;\theta_i), \ (\bm x,t)\in\Omega\times[0,T]
\end{equation}
for $\bc\in\mathbb R^N$. On the other hand, the input functions are mapped to the output latent space through the loss function \cref{eq:physics-l2-loss} in the optimization problem during the online phase. In contrast to aforementioned neural operators in \cref{sec:unified-ol-framework}, ReBaNO directly maps the input function to the latent representation of the solution by the online fine-tuning on the coefficient $\bc$ as the parameter of a single-layer neural network that minimizes the physics-informed loss \cref{eq:physics-l2-loss}, i.e.,
\begin{equation}\label{eq:rebono-encoder}
    \mathcal E_i^{\rm ReBaNO}:f\mapsto \bc, \text{ and } \mathcal{E}_i^{\rm ReBaNO}(f)\coloneq \mbox{\rm arg}\hspace*{-1pt}\min_{\bc \in\mathbb R^N} \mathcal{L}_p(\tilde{u};\bc)
\end{equation}
where the loss function is defined on a single input $f\in\mathcal F$. In summary, the target operator is approximated by
\begin{equation}
    \Psi^\sim(f;\theta)=\mathcal D_o^{\rm ReBaNO}\circ\mathcal{E}_i^{\rm ReBaNO}(f),
\end{equation}
where each component is defined in \cref{eq:rebano-decoder} and \cref{eq:rebono-encoder}, and the parameter $\theta$ collects the PINN parameters $\{\theta_i\in\Theta_{\rm PINN}\}_{i=1}^N$ along with the coefficients $\bc$.

\section{Numerical Results}\label{sec:results}

In this section, we compare the proposed ReBaNO with other data-driven learning approaches (PCA-Net, DeepONet, FNO and CNO) to approximating maps between infinite-dimensional function spaces using three benchmark problems. In \cref{subsec:poisson}, we start with the one-dimensional Poisson equation; \cref{subsec:darcyflow} studies the map from the permeability field to the solution for 2D steady Darcy flow; and in \cref{subsec:ns} we consider the Navier-Stokes equation. Because operators are expected to predict the outputs for the inputs in or out of the distribution seen in the training, we perform both in-distribution and out-of-distribution tests in our numerical experiments. The OOD datasets are generated from a similar measure from which the training data are sampled but with smaller correlation lengths. In this section, we focus on the discussion of the more challenging OOD tests. 
In the following numerical examples, unless otherwise specified, 1000 different input-output pairs are used for training data-driven models and 200 for testing. ReBaNO utilizes the same 1000 inputs as the discrete input data $F$ for greedy selection. We use a relative $L^2$ loss function for the training of data-driven models. Additional information for the experimental setup is given in \cref{sec:appdixA}. We emphasize here that during the sampling process of all examples, the absolute values of all random numbers are bounded by 4.  As a baseline example, to probe the benefit of physics constraints in improving the generalizability, we also contrast ReBaNO with PINO in Poisson's example. Due to the highly complex loss landscapes and the non-convexity of the optimization, PINO errors plateau above 20\% when applied to Darcy flow and Navier-Stokes problems. Therefore, the PINO results of Darcy and Navier-Stokes are not included here. All neural networks are trained on a single NVIDIA A100 using 32GB memory. For readers interested in the size and computational cost of each model, we refer to \cref{sec:suppl-figs}. The code for all numerical experiments can be found at \url{https://github.com/haolanzheng/rebano}.

\begin{table*}[htbp]
    \begin{small}
    \centering
    \caption{Poisson (top), Darcy flow (middle), and Navier-Stokes (bottom) benchmarks of all models. Note that PINO results are not included for Darcy and Navier-Stokes due to its errors plateauing above 20\% when applied to these two examples.}
    \begin{tabular}{ccccccccccc}
        \toprule 
        model & \# of params & $e_{\mathrm{mean}}^{\mathrm{train}}$ & $e_{\mathrm{max}}^{\mathrm{train}}$ & \multicolumn{2}{c}{$e_{\mathrm{mean}}^{\mathrm{test}}$} & \multicolumn{2}{c}{$e_{\mathrm{max}}^{\mathrm{test}}$} & \multicolumn{2}{c}{$e_{\rm mean}^{\rm test}/e_{\rm mean}^{\rm train}$} & Data\\
        &&&& ID & OOD & ID & OOD & ID & OOD & \\
        \midrule
        PCA-Net & \num{17670} & \num{0.003} & \num{0.051} & \num{0.011} & \num{0.100} & \num{0.310} & \num{0.767} & \num{3.965} & \num{35.919} & Yes \\
        DeepONet & \num{19021} & \num{0.003} & \num{0.028} & \num{0.011} & \num{0.087} & \num{0.451} & \num{0.933} & \num{3.811} & \num{30.511} & Yes \\
        FNO & \num{94657} & \num{0.002} & \textbf{\num{0.010}} & \num{0.003} & \num{0.015} & \num{0.040} & \num{0.138} & \num{1.564} & \num{6.976} & Yes \\
        CNO & \num{103989} & \num{0.002} & \num{0.011} & \num{0.005} & \num{0.052} & \num{0.134} & \num{1.414} & \num{1.162} & \num{23.941} & Yes \\
        PINO & \num{94657} & \num{0.002} & \num{0.014} & \num{0.002} & \textbf{\num{0.007}} & \num{0.075} & \num{0.095} & \num{1.199} & \textbf{\num{4.090}} & Yes \\
        ReBaNO & $8 + 901\times8$ & \textbf{\num{0.001}} & \num{0.021} & \textbf{\num{0.001}} & \num{0.008} & \textbf{\num{0.023}} & \textbf{\num{0.071}} & \textbf{\num{1.001}} & \num{6.975} & \textbf{No} \\
        
        \bottomrule
        \midrule
        PCA-Net & \num{410733} & \num{0.025} & \num{0.064} & \num{0.057} & \num{0.065} & \num{0.167} & \num{0.157} & \num{2.309} & \num{2.634}  & Yes \\
        DeepONet & \num{384601} & \num{0.018} & \num{0.038} & \num{0.061} & \num{0.069} & \num{0.165} & \num{0.223} & \num{3.419} & \num{3.899} & Yes \\
        FNO & \num{412929} & \num{0.020} & \num{0.085} & \num{0.032} & \num{0.038} & \num{0.089} & \num{0.083} & \num{1.545} & \num{1.876} & Yes \\
        CNO & \num{359877} & \textbf{\num{0.004}} & \textbf{\num{0.011}} & \textbf{\num{0.010}} & \textbf{\num{0.013}} & \textbf{\num{0.040}} & \textbf{\num{0.038}} & \num{2.489} & \num{3.247} & Yes \\
        ReBaNO & $48 + 8361\times48$ & \num{0.045} & \num{0.107} & \num{0.044} & \num{0.048} & \num{0.100} & \num{0.104} & \textbf{\num{0.976}} & \textbf{\num{1.085}} & \textbf{No} \\
        \bottomrule
        \midrule
        PCA-Net & \num{86221} & \num{0.052} & \num{0.961} & \num{0.317} & \num{0.859} & \num{0.959} & \num{2.578} & \num{6.097} & \num{16.349} & Yes  \\
        DeepONet & \num{51601} & \num{0.047} & \num{0.731} & \num{0.379} & \num{1.315} & \num{1.610} & \num{5.514} & \num{7.996} & \num{27.755} & Yes \\
        FNO & \num{58961} & \textbf{\num{0.002}} & \textbf{\num{0.005}} & \textbf{\num{0.002}} & \textbf{\num{0.006}} & \textbf{\num{0.005}} & \textbf{\num{0.013}} & \num{1.018} & \num{3.781} & Yes \\ 
        CNO & \num{93783} & \num{0.004} & \num{0.008} & \num{0.005} & \num{0.021} & \num{0.008} & \num{0.044} &\num{1.384} & \num{5.690} & Yes \\
        ReBaNO & $20 + 2222\times20$ & \num{0.036} & \num{0.069} & \num{0.036} & \num{0.056} & \num{0.072} & \num{0.135} & \textbf{\num{0.996}} & \textbf{\num{1.569}} & \textbf{No} \\
        \bottomrule
    \end{tabular}
    \label{tab:all-benchmarks} \\
    \vspace{4pt}
    *The last column indicates whether high-fidelity solutions are needed for training the models. The boldface highlights the best performer of each benchmark. 
\end{small}
\end{table*}
In our numerical experiments, we use $L^2$ relative error to measure the accuracy of the predictions given by all models
\begin{equation}
    e=\frac{\|u_{\rm pred} - u\|_{L^2}}{\|u\|_{L^2}}
\end{equation}
where $u_{\rm pred}$ is the prediction and $u$ the high-fidelity solution. Meanwhile, to measure the generalizability, we propose a metric of the ratio of the mean relative error on test datasets to that on training datasets
\begin{equation}
    r = \frac{e^{\rm test}_{\rm mean}}{e^{\rm train}_{\rm mean}}
\end{equation}
A model has good generalizability if the ratio is close to 1. \par 

For an overview of the performance of all models, the benchmarks of three examples are listed in the \cref{tab:all-benchmarks}. Based on the results of these benchmark problems, ReBaNO exhibits significantly better generalizability even to both ID and OOD datasets compared to all other data-driven models. More importantly, as shown below, the accuracy of ReBaNO shows strict discretization invariance, while the accuracy of FNO and CNO significantly deteriorates when the grid size deviates from where FNO and CNO are trained.

\subsection{1D Poisson Equation}\label{subsec:poisson}

\begin{figure*}[htbp]
    \centering
        \includegraphics[width=0.9\linewidth]{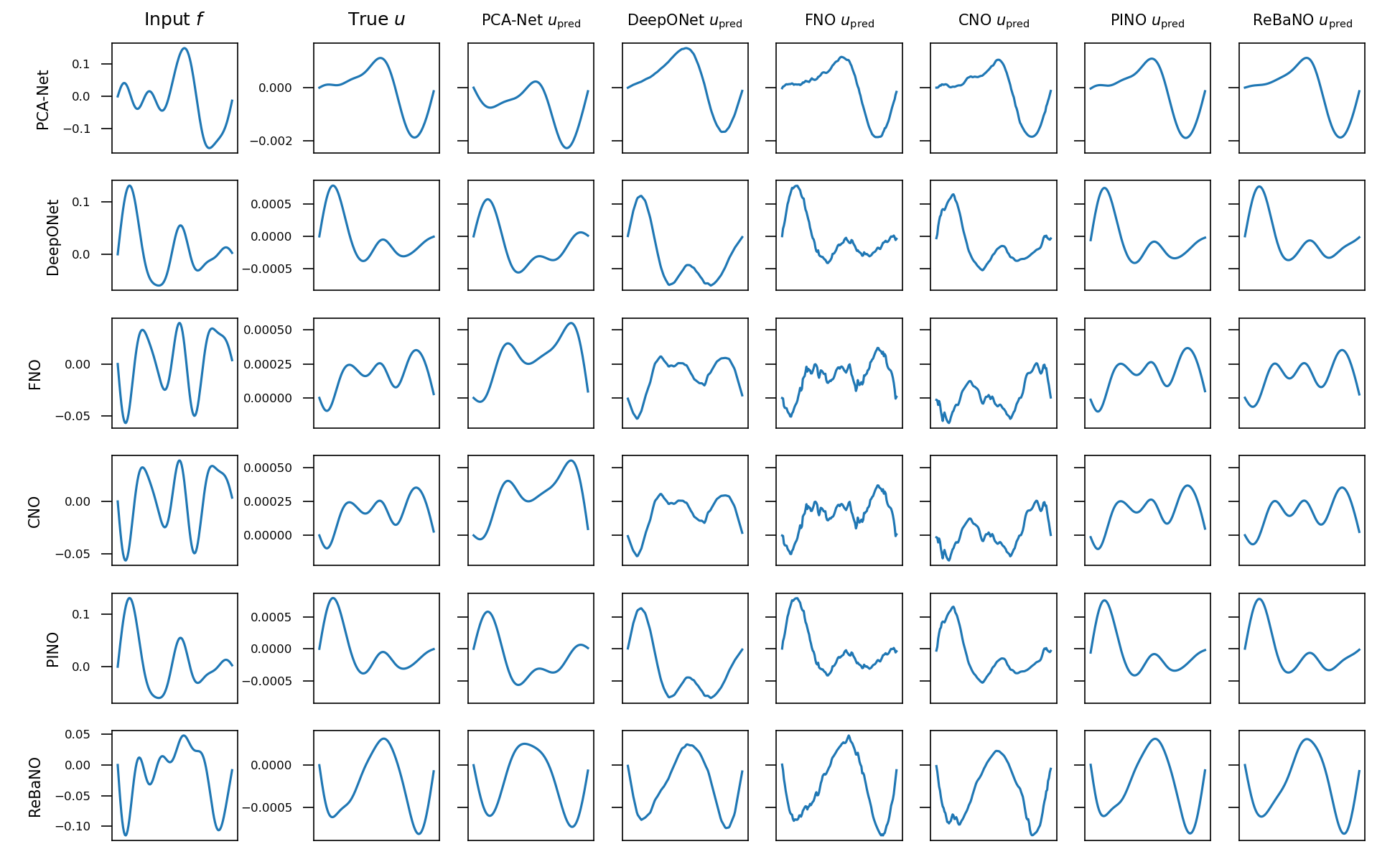}
    \caption{Poisson: inputs and true outputs in the OOD test resulted in worst case errors for each model. PCA-Net is composed of six hidden layers and 64 neurons per layer. DeepONet consists of a branch net with four hidden layers of width 60 and a trunk net with three hidden layers of width 60. FNO is composed of one lifting layer, one projection layer and two Fourier neural layers with 64 neurons for each layer. CNO employs 4 up/down sampling blocks, two residual blocks and another two in the neck. PINO uses the same architecture as the FNO. 8 pre-trained full PINNs are used in the implementation of our ReBaNO.}
    \label{fig:poisson1d-worst-test-cases-ood}
\end{figure*}
As the first numerical example, we consider 1D Poisson equation define on the interval $\Omega=(0,1)$
\begin{align} 
-&\dv[2]{u}{x}=f(x), \ x\in\Omega \label{eq:poisson-eq}\\
&u(0)=u(1)=0 \label{eq:poisson-bc} 
\end{align}
The data of the source term $f(x)$ are sampled from a centered Gaussian measure $N(0, C)$ with covariance $C=(-\mathrm{d}^2/\mathrm{d}x^2+1)^{-2}$, where $-\mathrm{d}^2/\mathrm{d}x^2$ is the Laplacian defined on $\Omega$ with homogeneous Dirichlet boundary condition. The data for OOD test are generated from a similar Gaussian with different covariance $C=(-\mathrm{d}^2/\mathrm{d}x^2+25)^{-2}$. The solutions to the Poisson equation are obtained analytically. We aim to learn a map from $f(x)$ to the solution $u(x)$: $\Psi:f\mapsto u$.  \par

\paragraph{Accuracy Test}

\begin{figure}[htbp]
    \centering
        \includegraphics[width=0.8\linewidth]{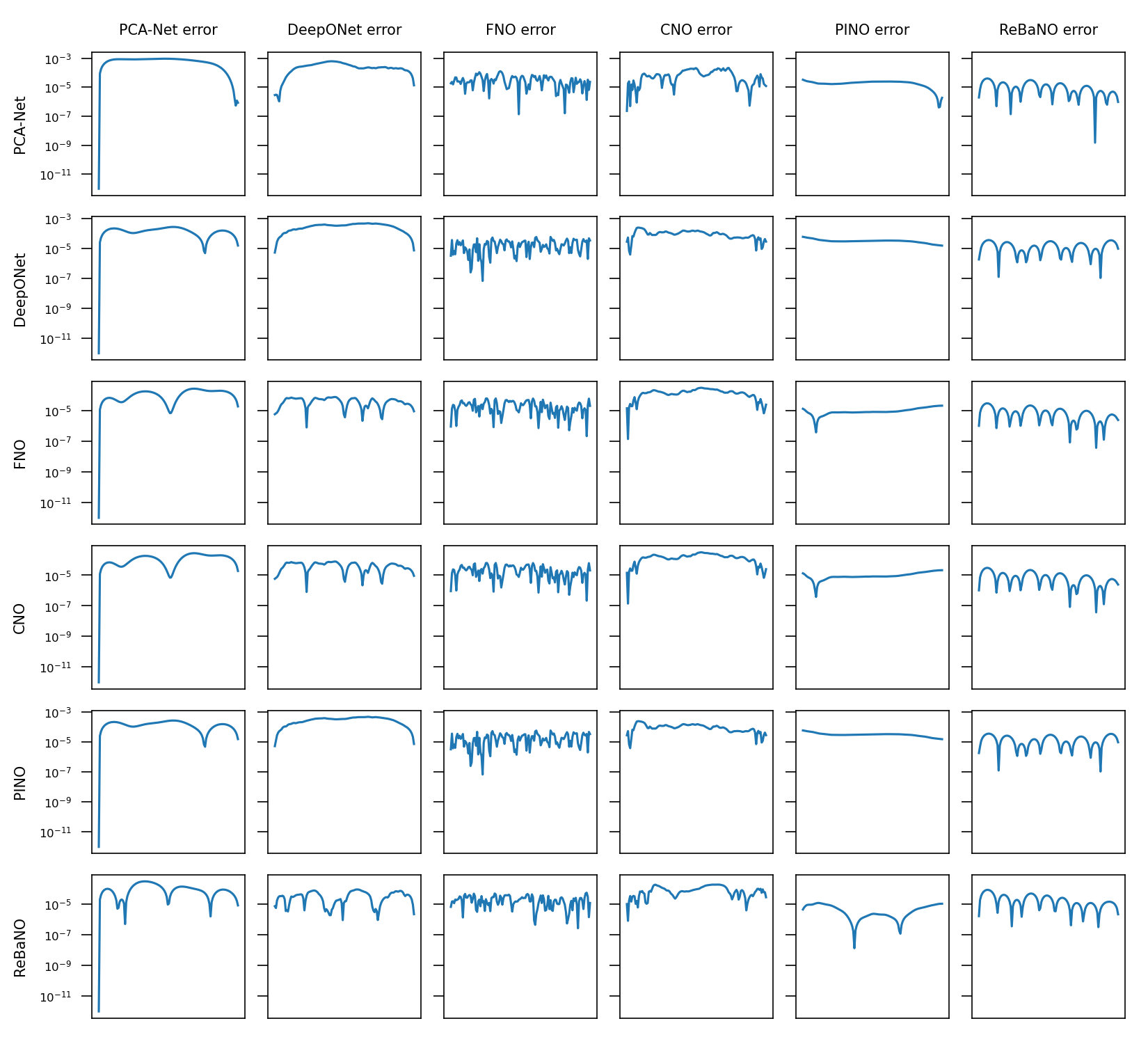}
    \includegraphics[width=0.8\linewidth]{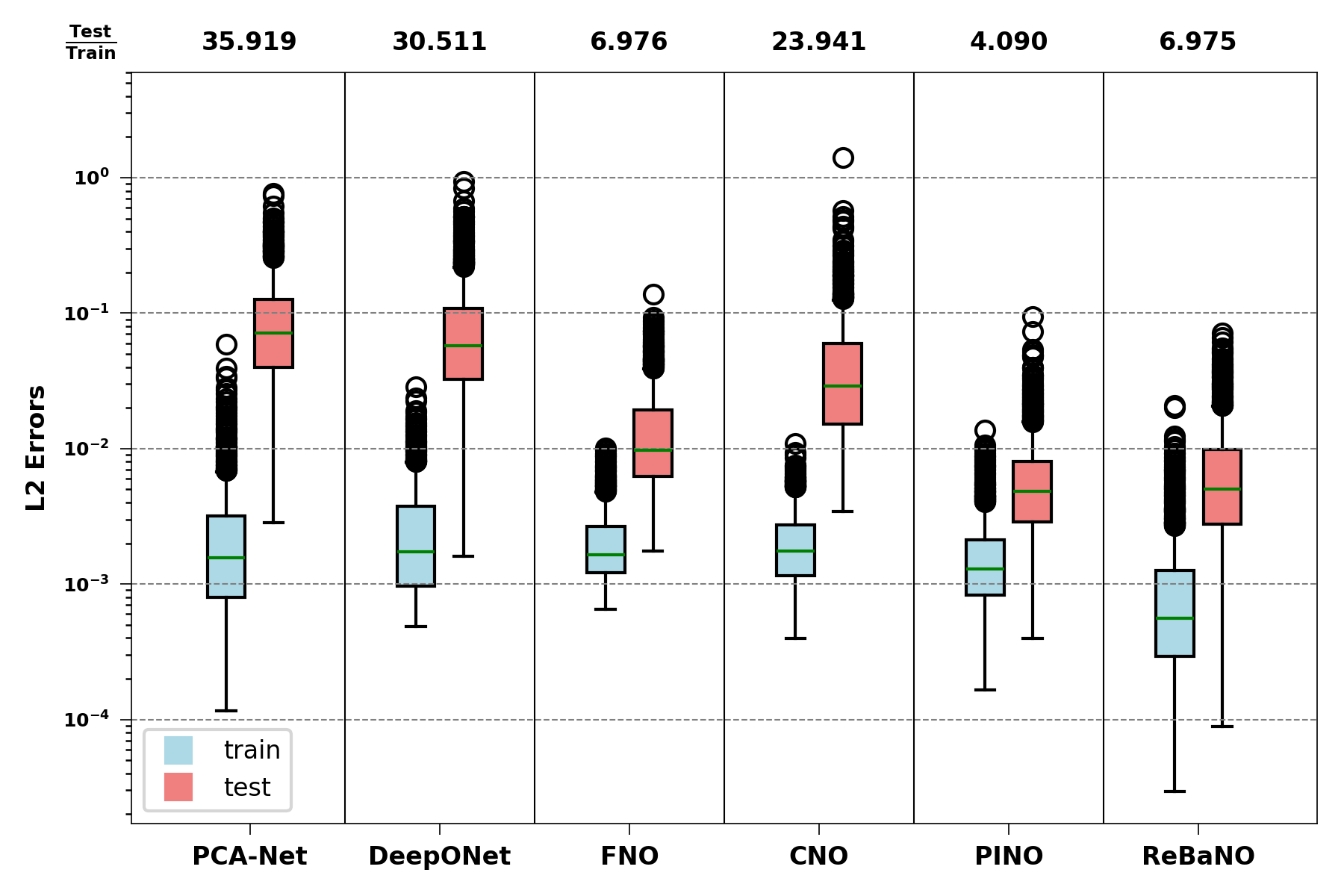}    
    \caption{Poisson OOD test pointwise (top) and $L^2$ relative (bottom) errors. Each subfigure on the top part in each row presents pointwise errors given by the six models for the cases that result in the worst-test-error cases of PCA-Net, DeepONet, FNO, CNO, PINO, and ReBaNO (from top to bottom). Shown on the bottom are the box plots of the $L^2$ relative errors of the six models. The numbers printed out are the ratios between mean test errors and mean training errors, measuring the generalizability of each model.}
    \label{fig:poisson1d-test-errors-comparison-ood}
\end{figure}

\cref{fig:poisson1d-worst-test-cases-ood} shows the inputs, true outputs, and predicted outputs given by six models for the inputs from the OOD dataset that result in the worst test errors of each model. For this simple 1D problem, the models succeed in capturing the main features of the solutions. PCA-Net and DeepONet give relatively poor accuracy compared to other models. To make a reasonable comparison, we present the pointwise errors in the OOD test on the worst-test-error case of each model in \cref{fig:poisson1d-test-errors-comparison-ood} Top. For example, the first row gives six graphs of pointwise errors given by PCA-Net, DeepONet, FNO, CNO, PINO and ReBaNO (from left to right), respectively of the case that results in the worst test error given by PCA-Net. 
We can see that, comparing PINO and ReBaNO to FNO, the physical constraint smooths the pointwise error. For the same input, FNO, CNO, PINO, and ReBaNO achieve similarly small test errors compared to PCA-Net and DeepONet.

In \cref{fig:poisson1d-test-errors-comparison-ood} Bottom, we also show the box plot of relative $L^2$ errors. The numbers printed on the top of each column are the ratios of the mean test errors to the mean training errors provided by all models. The benchmarks of the six models on the Poisson problem are listed in \cref{tab:all-benchmarks} (top). From the table and \cref{fig:poisson1d-test-errors-comparison-ood} Bottom, among all purely data-driven models, FNO gives the most consistent level of accuracy in the OOD test. Compared to ID tests, there are wider generalization gaps in all models, while physics-informed models outperform all data-driven models in terms of generalizability. And PCA-Net and DeepONet generalize poorly to OOD data, where the test errors are 30 times more than the training errors. Among all models, PINO achieves the smallest ratio in the OOD test while ReBaNO yields the smallest errors.

\paragraph{Ablation Study on Neuron Selection}

\begin{figure}[htbp]
    \centering
    \includegraphics[width=0.9\linewidth]{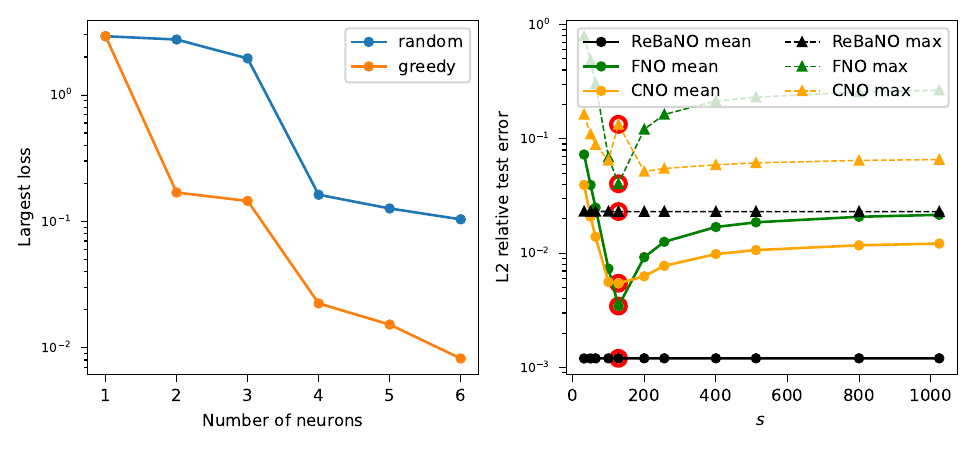}
    \caption{Left: Poisson ablation test on greedy algorithm, showing the largest loss as the number of neurons, selected with or without greedy algorithm, increases. Right: Discretization Invariance Test. Shown are the variations of the mean and max test errors of FNO, CNO and ReBaNO with respect to different discretizations (the number of grid points varies as $s=32,50,64,100,128,200,256,400,512,800,1024$). The large red dots denote that all models are trained on the grid of size 128.}
    \label{fig:ablation-discretization}
\end{figure}

To see the importance of greedy selection, we randomly sample six neurons (i.e., the set $\{\Psi_h(f_i)\}_{i=1}^6$ for random source terms $\{f_i(x)\}_{i=1}^6$) from the same input data. From \cref{fig:ablation-discretization} Left, it is clear that with neurons selected using greedy algorithm the largest loss of ReBaNO is more than 10 times smaller than that with randomly selected neurons. 

\paragraph{Discretization Invariance Test}

As discussed in previous work (see Refs. \cite{li2021fourier, kovachki2023neural}), it is important for a robust operator learning model to be invariant with respect to different discretizations or resolutions. In \cref{fig:ablation-discretization} Right, we compare the test errors of FNO, CNO, and ReBaNO when varying the grid size on which the input functions are evaluated. Specifically, all models are trained on a grid of size 128. The test inputs for $f(x)$ are sampled from the same Gaussian measure as for the training data, but with different numbers of grid points $s=32,50,64,100,128,200,256,400,512,800,1024$. As mentioned in \cref{subsec:olfw-olfw}, DeepONet and PCA-Net both depend on input resolution. Therefore, we do not compare with DeepONet and PCA-Net. From \cref{fig:ablation-discretization} Right, we show the mean and maximum relative $L^2$ error of FNO, CNO, and ReBaNO. We see that the mean and maximum errors given by FNO and CNO increase significantly in super/sub-resolution tests. In the sub-resolution regime, the error is amplified at most more than 10 times, and is increased with a factor of 5 in the super-resolution regime. ReBaNO presents highly consistent accuracy across all resolutions, implying that our ReBaNO successfully learns a discretized approximator that preserves continuous-discrete-equivalence to the ground-truth operator. In low-resolution regime, the errors of FNO and CNO are dominant by the aliasing error arising from the discretizations of $f(x)$. In the high-resolution regime, the frequency truncation mainly contributes to the errors. Note that the maximum error of the CNO reaches a local maximum at $s=128$ where it is trained. This is because the CNO is trained with the mean $L^2$ loss, not the sup-norm, and the CNO up/downsamples the inputs to the fixed input grid, meaning that the CNO may overlook the discrepancies when dealing with coarser inputs, and the errors are also smoothed with higher resolutions. The reason why ReBaNO presents strong discretization invariance is that PINNs learn continuous maps from the physical domain to solution values in a mesh-free manner instead of dealing with mappings between functions, leading to strictly consistent accuracies of the ReBaNO across all discretizations.

\subsection{2D Steady Darcy Flow}\label{subsec:darcyflow}
    Next we consider two-dimensional steady-state Darcy flow problem defined on the unit square $\Omega=[0,1]^2$.
    \begin{align}
        -\nabla\cdot(a(\bm{x})\nabla u(\bm{x}))&=f(\bm{x}), \ \bm{x}\in\Omega \label{eq:darcy-eq} \\
        u(\bm{x})&=0, \ \bm{x}\in\partial\Omega \label{eq:darcy-bc}
    \end{align} 
    where $a\in L^\infty(\Omega;\mathbb{R}_+)$ is the permeability field. The force term $f(\bm{x})=1$ is fixed. Following \cite{li2021fourier}, the function $a(\bm{x})$ is sampled from the measure $\mu=T_\#N(0,C)$ as the pushforward of a Gaussian measure under the operator $T$ with the covariance $C=(-\Delta+9)^{-2}$, where $-\Delta$ is the Laplacian defined on $\Omega$ and a homogeneous Neumann boundary condition is imposed. The operator $T$ is defined in a piecewise fashion as 
    
    \begin{equation}
        T(x)=\left\{ 
        \begin{aligned}
            &12, \quad x\geq 0\\
            &3, \quad x<0.
        \end{aligned}\right.
    \end{equation}
    The OOD data are sampled from a similar Gaussian with a covariance $C=(-\Delta+64)^{-2}$. High-fidelity solutions are computed using FEM over a $100\times100$ grid. We seek to learn a map from the permeability field $a(\bm{x})$ to the solution $u(\bm{x})$: $\Psi:a\mapsto u$. \par 
    
    \begin{figure*}[htbp]
        \centering
            \includegraphics[width=0.9\linewidth]{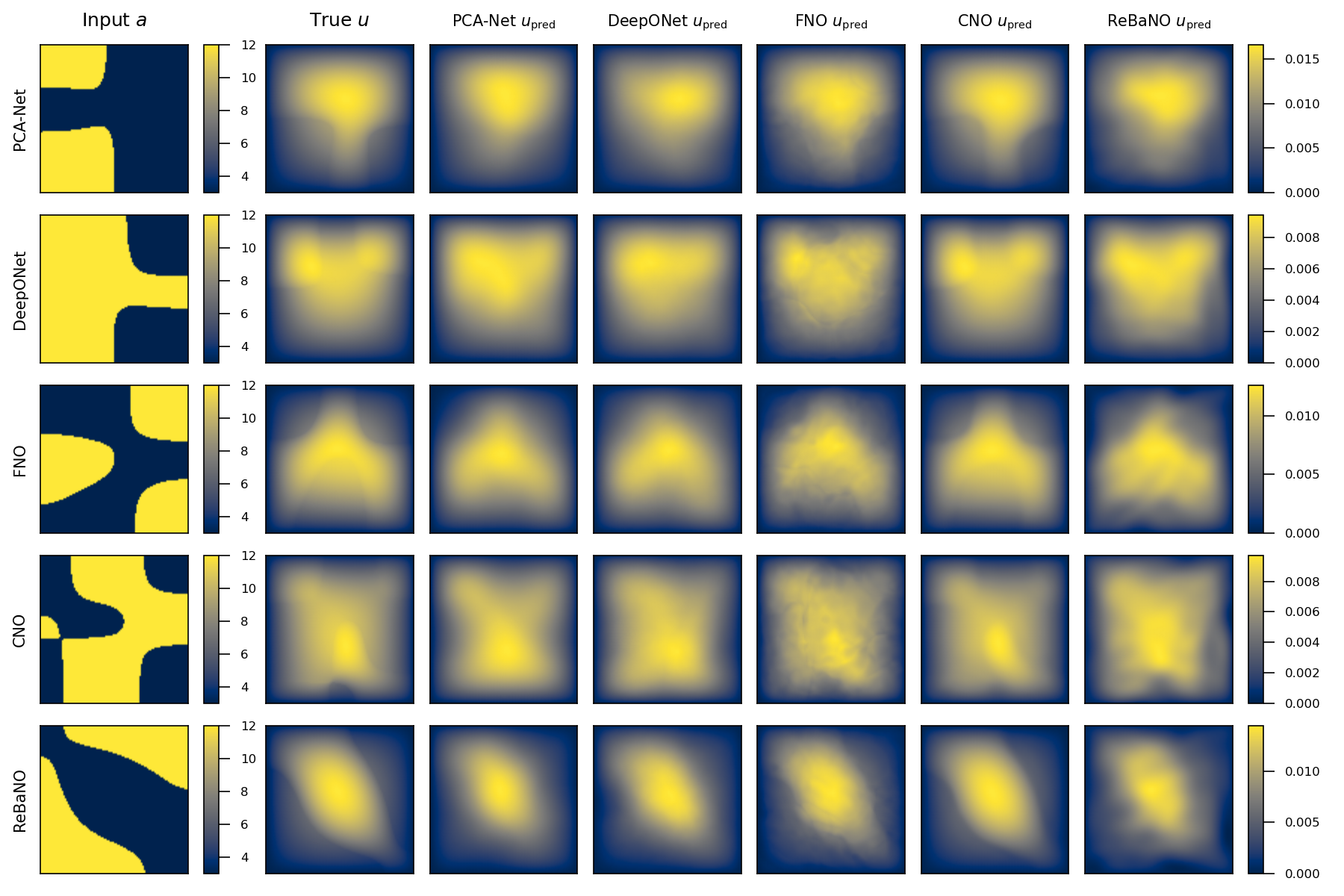}
        \caption{Darcy flow: inputs and true outputs in OOD test that result in worst case errors. PCA-Net is composed of five hidden layers and 200 neurons per layer and DeepONet consists of a branch net with five hidden layers of width 200 and a trunk net with four hidden layers of the same width. FNO is composed of one lifting layer, one projection layer and two Fourier neural layers with 32 neurons for each layer. CNO is composed of four up/down sampling blocks, four residual blocks and two residual blocks in the neck. 48 pre-trained RVPINNs are used in the implementation of our ReBaNO.}
        \label{fig:darcy-worst-test-cases-ood}
    \end{figure*}

\begin{figure}[htbp]
        \centering
            \includegraphics[width=0.75\linewidth]{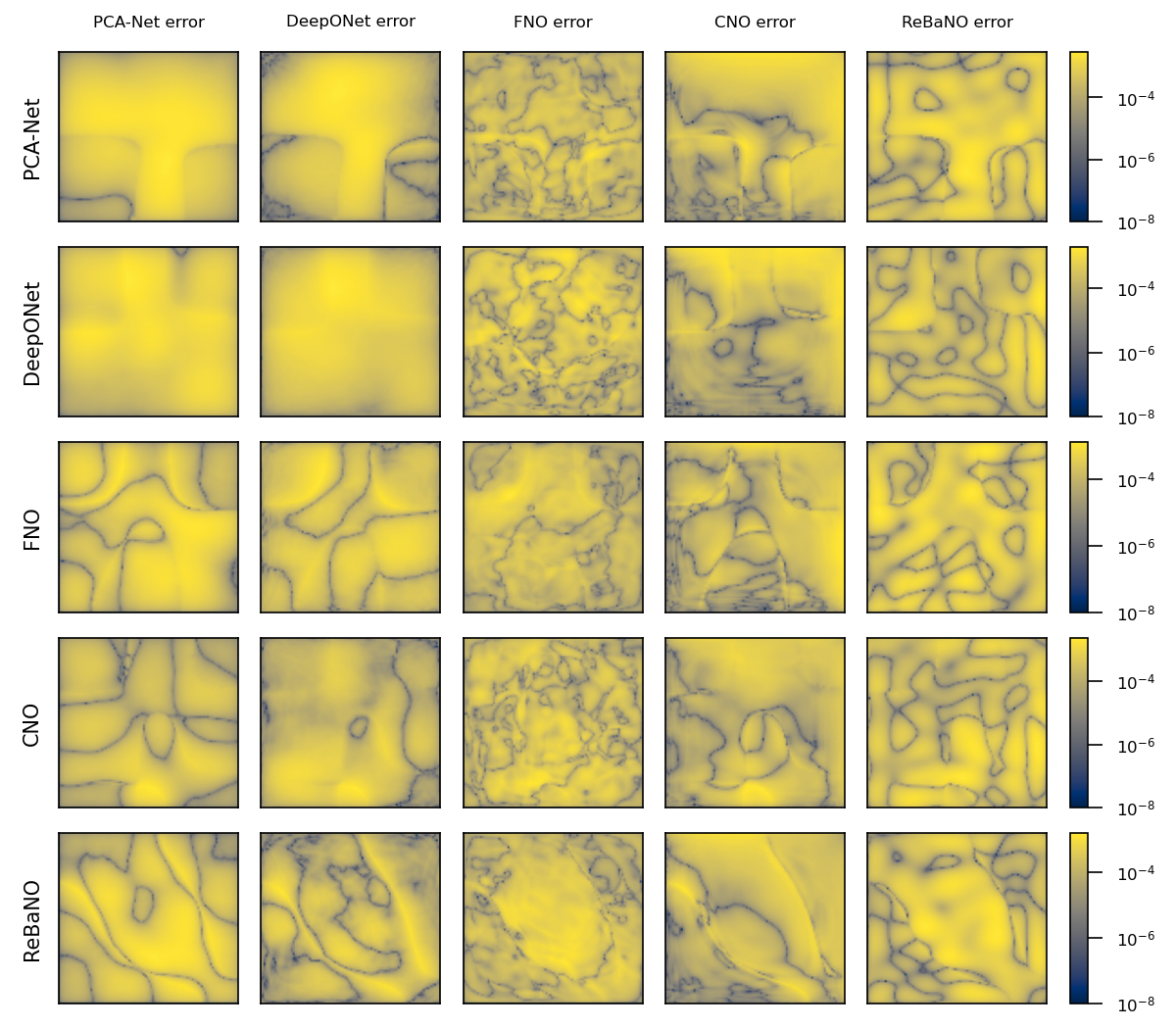}
        \includegraphics[width=0.8\linewidth]{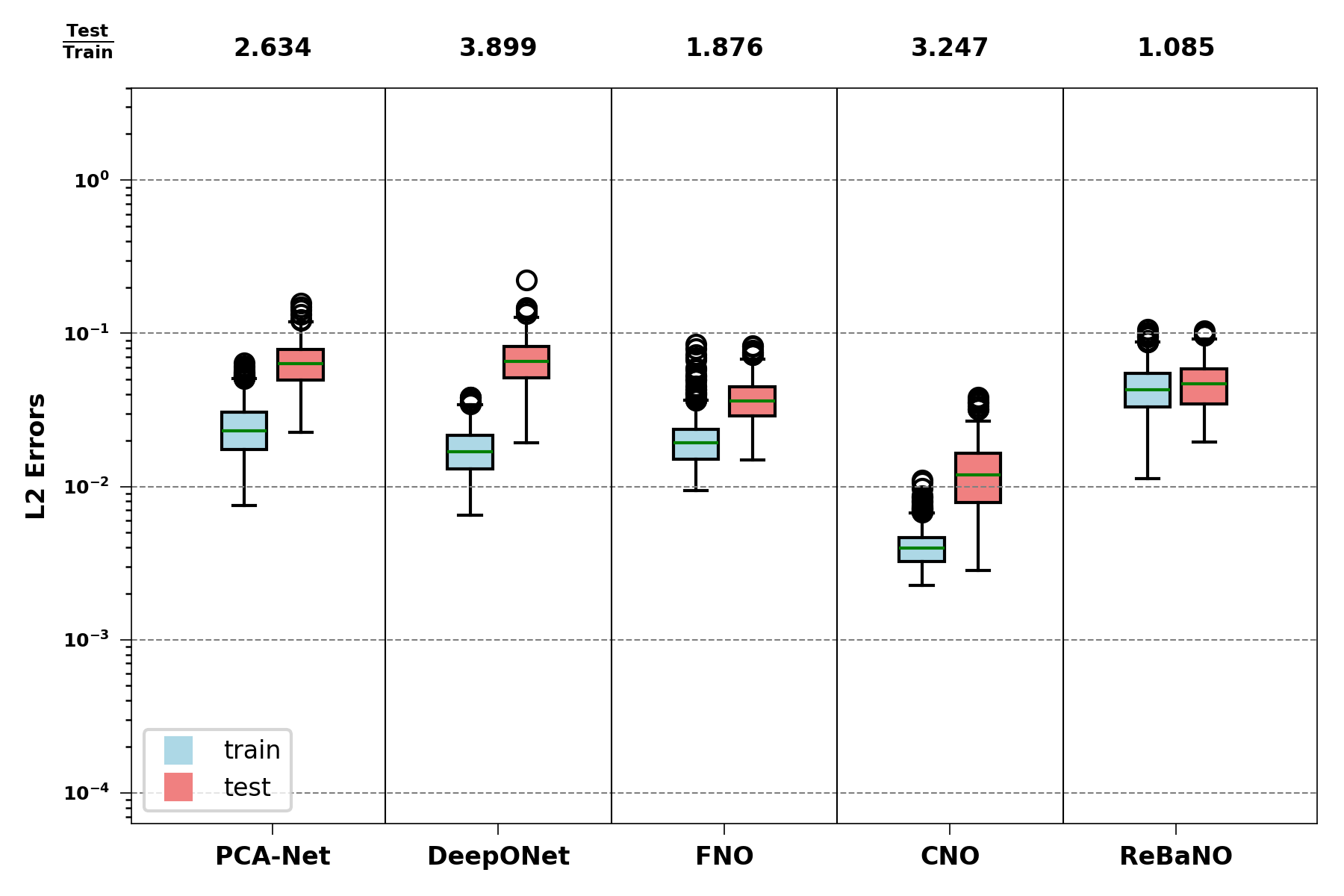}
        \caption{Darcy flow OOD test pointwise (top) and $L^2$ relative (bottom) errors. Each subfigure on the top part in each row presents pointwise errors given by the five models on the cases that result in the worst-test-error cases of PCA-Net, DeepONet, FNO, CNO and ReBaNO (from top to bottom). Shown on the bottom are the box plots of the  $L^2$ relative errors of the five models. The numbers printed out are the ratios between mean test errors and mean training errors, measuring the generalizability of each model.}
        \label{fig:darcy-test-errors-comparison-ood}
    \end{figure}

\paragraph{Accuracy Test}    Similarly to the 1D Poisson example, \cref{fig:darcy-worst-test-cases-ood} shows the input, true output, and predicted output given by five models for the inputs from OOD data that result in worst test errors of each model. All models succeed in capturing the main features of the solution, where CNO achieves the best accuracy. From the similar comparison of pointwise errors in worst-test-error cases presented in \cref{fig:darcy-test-errors-comparison-ood} Top, we find that FNO predictions exhibit more explicit granularity.     The box plot of $L^2$ relative errors in OOD test is outlined in \cref{fig:darcy-test-errors-comparison-ood} Bottom and benchmarks for the Darcy flow problem are listed in \cref{tab:all-benchmarks} (middle). \cref{fig:darcy-test-errors-comparison-ood} Bottom and \cref{tab:all-benchmarks} (middle) also indicate that ReBaNO presents the most comparable performance between the training data and the OOD test data, while the error of all data-driven models increases by a factor of 2 or 3.

\subsection{2D Navier-Stokes Equation}\label{subsec:ns}

    In this subsection, we study the two-dimensional Navier-Stokes equation. 
        \begin{align}
            \pdv{\omega}{t}+\bm{u}\cdot\nabla\omega &= f^\prime + \nu\Delta \omega,\  (\bm{x},t)\in \Omega\times[0, T]\label{eq:nse}\\
            \omega &= -\Delta \psi \label{eq:nse-vor-stream} \\
            \omega(\bm{x},0) &= \omega_0(\bm{x}) \label{eq:nse-ic}, \ \bm{x}\in \Omega
        \end{align}
    \begin{figure}[H]
        \centering
            \includegraphics[width=0.9\linewidth]{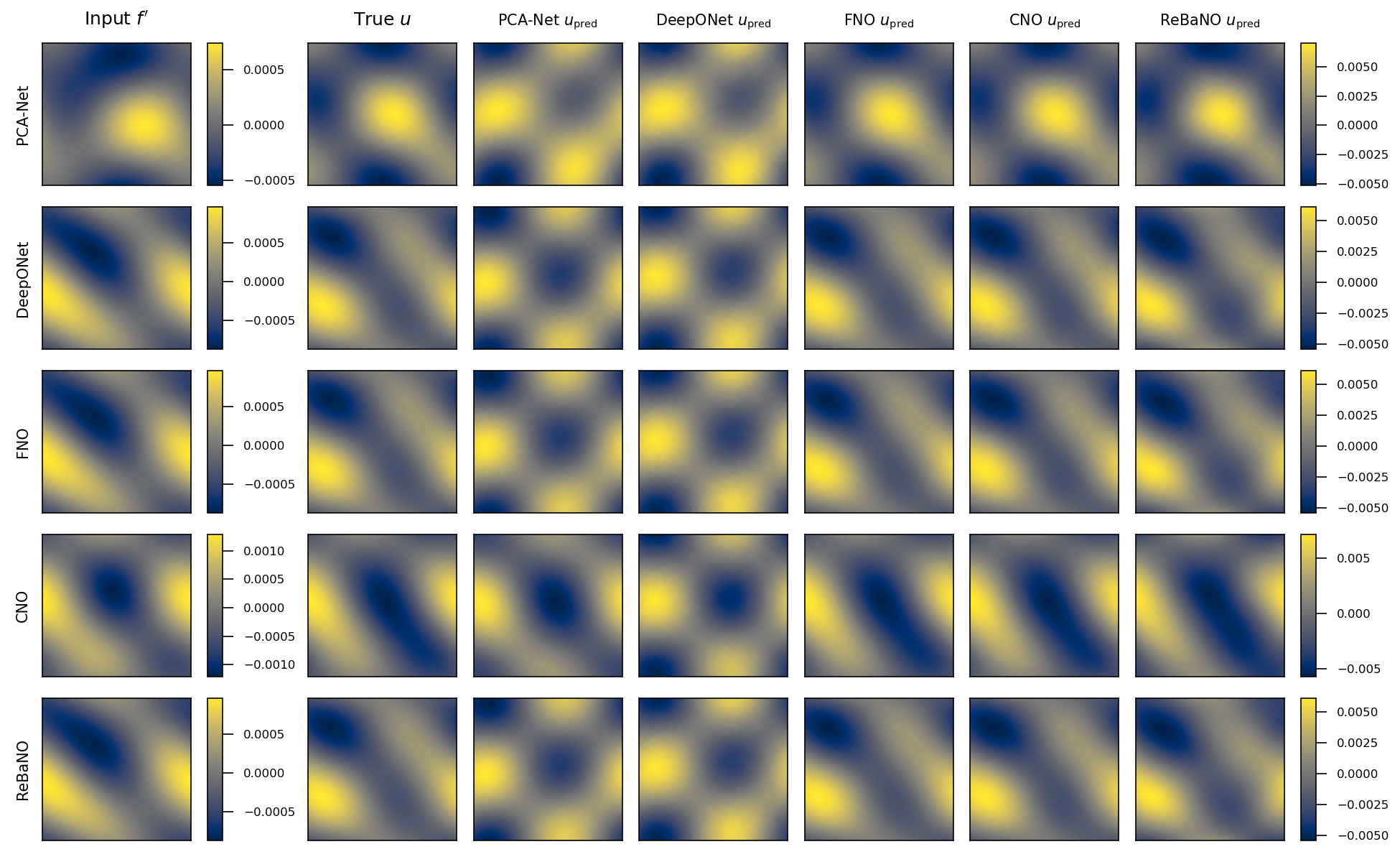} 
        \caption{Navier-Stokes: inputs and true outputs of the vorticity resulted in worst case test errors. PCA-Net is composed of four layers and 200 neurons per layer and DeepONet consists of a branch net with four layers of width 100 and a trunk net with three layers of the same width. FNO is composed of one lifting layer, one projection layer and two Fourier neural layers with 18 neurons for each layer. CNO uses four up/down sampling blocks and four residual blocks inside along with two residual blocks in the neck. 20 pre-trained PINNs are used in the implementation of our ReBaNO.}
        \label{fig:ns-worst-test-cases-ood}
    \end{figure}
    Here we consider a square domain $\Omega=[0, 2\pi]^2$ with periodic boundary conditions, where $\bm{u}$ is the physical velocity field and it is related to $\psi$ by $\bm{u}=(\partial_y \psi, -\partial_x \psi)$. And $f^\prime(\bm{x})$ is the curl of the external force and $\nu$ the viscosity constant. We aim to learn the map from $f^\prime$ to the vorticity field $\omega$ at a fixed moment $T=10$: $\Psi: f^\prime(\bm{x}) \mapsto \omega(\bm{x},T)$. We sample the data of $f^\prime$ from a centered Gaussian with covariance $C=(-\Delta+9)^{-4}$, where the Laplacian $-\Delta$ is defined on $\Omega$ subject to a periodic boundary condition. The initial condition $\omega_0$ is fixed and sampled from the same measure. OOD data are samples from a similar measure with $C=(-\Delta+25)^{-4}$. Note that \cref{eq:nse,eq:nse-vor-stream,eq:nse-ic} and periodic boundary conditions can only determine the stream function $\psi$ up to an arbitrary additive constant. We can impose an additional condition to eliminate arbitrariness. In this work, we enforce $\psi$ to be a mean-free function over the spatial domain $\Omega$:
    \begin{equation}
    \int_\Omega\psi = 0
    \end{equation}
    We use pseudo-spectral method with a tiny time step size ($\Delta t=1/5000$) to solve the Navier-Stokes equation over a $100\times100$ grid.

\paragraph{Accuracy Test} 

\begin{figure}[htbp]
    \centering
        \includegraphics[width=0.75\linewidth]{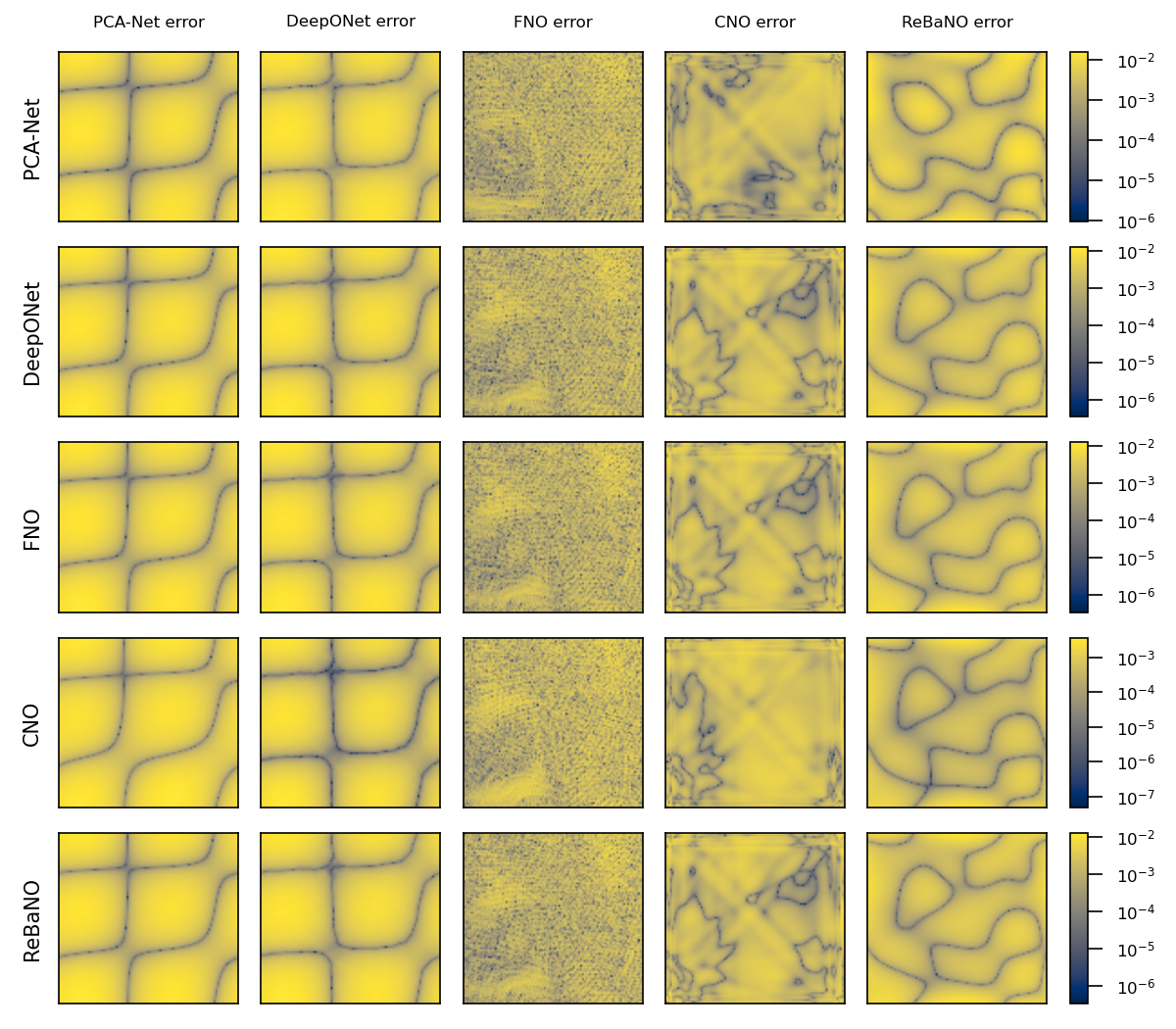}
    \includegraphics[width=0.8\linewidth]{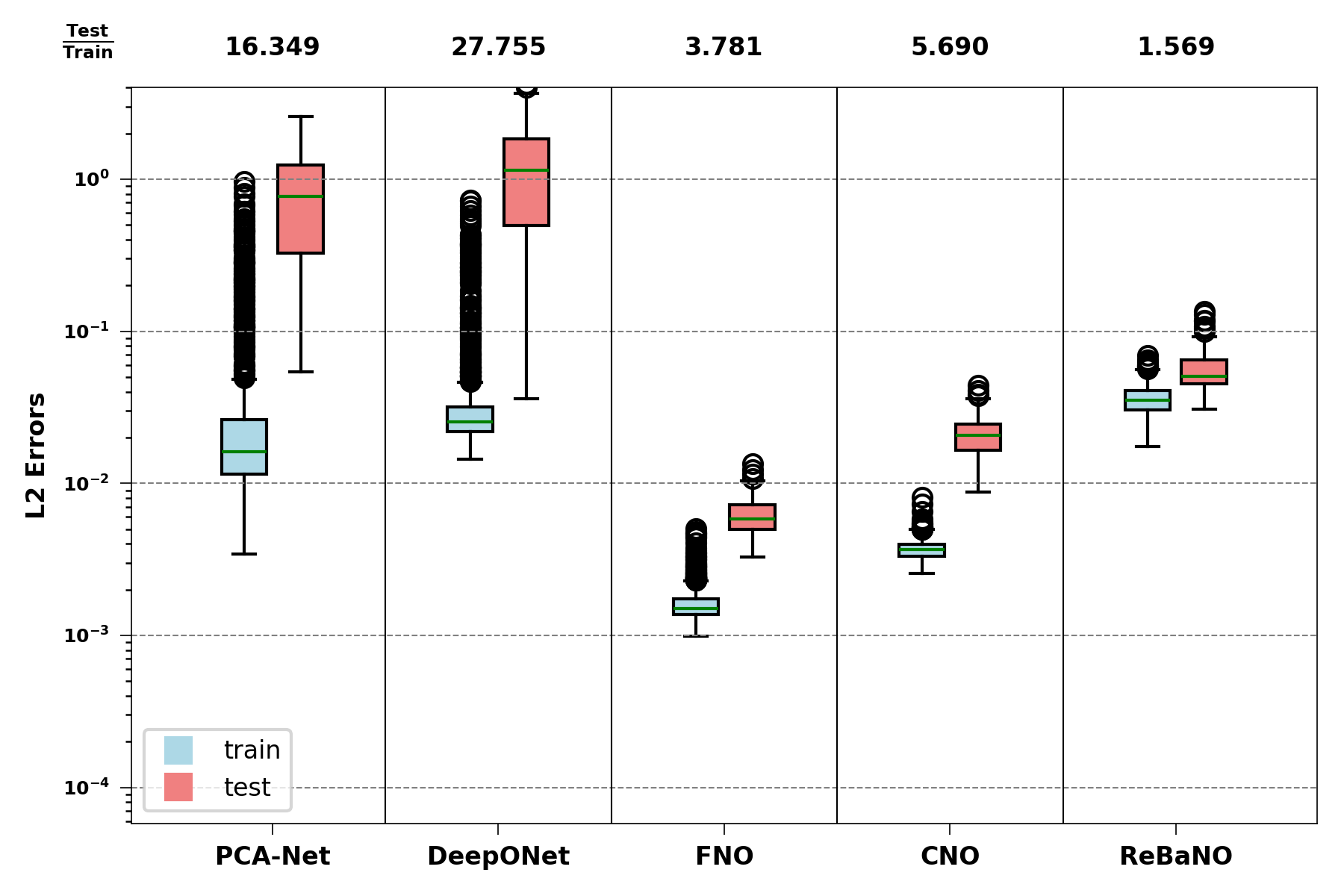}
    \caption{Navier-Stokes OOD test pointwise (top) and $L^2$ relative (bottom) errors. Each subfigure on the top part in each row presents pointwise errors given by the models for the cases that result in the worst-test-error cases of PCA-Net, DeepONet, FNO, CNO and ReBaNO (from top to bottom). Shown on the bottom are the box plots of the L2 relative errors of the five models. The numbers printed out are the ratios between mean $L^2$ relative test errors and mean $L^2$ relative train errors, measuring the generalizability of each model.}
    \label{fig:ns-test-errors-comparison-ood}
\end{figure}

    In \cref{fig:ns-worst-test-cases-ood}, it shows the input, true vorticity and predicted vorticity given by the five models for the inputs from OOD data that lead to the worst test errors of each model. A similar pointwise error comparison plot is in \cref{fig:ns-test-errors-comparison-ood} Top. For this problem, the vorticity field is well-correlated with the external force. DeepONet performs worst on its worst-test-error case. \par

    The box plot of $L^2$ relative errors for vorticity in the OOD test is plotted in \cref{fig:ns-test-errors-comparison-ood} Bottom and the corresponding benchmarks are listed in \cref{tab:all-benchmarks} (bottom). PCA-Net and DeepONet show markedly lower accuracy and generalizability than the other models on OOD inputs. When extrapolating to OOD data, FNO and CNO errors increase by a factor of 3 and 5, respectively, whereas ReBaNO error increases only by approximately 50\%.

\section{Conclusion}\label{sec:conclusion}

This work advances a reduced basis-driven physics-informed operator learning approach. We have introduced a data-lean operator learning algorithm, the Reduced Basis Neural Operator (ReBaNO). Guided by a mathematically rigorous greedy algorithm and with the notion of embedding physics, ReBaNO leverages knowledge distillation via task-specific activation function. After an offline-training stage which relies on minimal amount of data, ReBaNO's online solver is extremely lightweight, with a network consisting of only one hidden layer in which each neuron is a full pre-trained PINN. 
Tests against four state-of-the-art operator learning models reveal the fact that ReBaNO is the only NO achieving full discretization invariance, and ReBaNO's capability in significantly shrinking the generalization gap for both in-distribution and out-of-distribution tests. Future work includes theoretical convergence rate analysis and augmentation toward robust OOD performance when boundary conditions are concerned. Both are extremely challenging topics in the model reduction and scientific machine learning communities.

\bibliographystyle{plain}

\appendix
\section{Experimental setup}
\label{sec:appdixA}
\subsection{Poisson Setup}
The full PINNs are trained on grid of size 128 uniformly discretized over $\Omega=[0,1]$ through 40000 epochs with a fixed learning rate of 0.0005. The architecture of each PINN is $[1,20,20,20,1]$ and the activation function is $\tanh$. And 100 epochs are used for both training and testing for fine tuning with a fixed learning rate of 0.8 using L-BFGS optimizer. Unless otherwise specified, the optimizer in all training and testing processes of ReBaNO is Adam. ReBaNO leverages 8 pre-trained neurons for this problem. As for data-driven models, PCA-Net has six hidden layers of width 64 and DeepONet consists of a branch net with four hidden layers of width 60 and a trunk net with three hidden layers of the same width. FNO is composed of two Fourier neural layers along with one lifting layer and one projection layer with 64 neurons per layer, and the maximum number of Fourier modes is set to be 10. PINO takes the same architecture as FNO. CNO consists of four down/up sampling blocks, two residual blocks in the middle, and two residual blocks in the neck. The channel multiplier is 8. All data-driven models are trained using Adam optimizer through 1000 epochs with an initial learning rate of 0.001 which is halved every 100 epochs. PINO is trained through 10000 epochs with the same learning rate halved every 1000 epochs. The activation function ReLU is employed. 1000 instances of $f(x)$ are used for testing in all models. 

\subsection{Darcy flow Setup}
    For Darcy flow, Robust Variational PINN learns the solution in a more appropriate manner because it leverages the weak-form loss function. We train RVPINN using $8\times8$ elements with $20\times20$ Gauss quadrature points for each element through 60000 epochs with an initial learning rate of 0.001 and it is halved after every 10000 epochs. The test functions we use are the tensor product of one-dimensional FEM piecewise linear functions. In the greedy search, the fine-tuning is carried over $8\times8$ elements with $16\times16$ Gauss quadrature points for each using the L-BFGS optimizer through 100 epochs for both training and testing with a fixed learning rate of 0.8. The architecture of each RVPINN is $[2,40,40,40,40,40,40,1]$ and the activation function is $\sin$. 48 neurons are trained for Darcy flow. Meanwhile, PCA-Net has five hidden layers of width 200 and DeepONet consists of a branch net with five hidden layers of width 200 and a trunk net with four hidden layers of the same width. FNO is composed of two Fourier neural layers along with one lifting layer and one projection layer with 32 neurons per layer, and the maximum number of Fourier modes is set to be 10. CNO consists of four down/up sampling blocks, four residual blocks in the middle, and two residual blocks in the neck. The channel multiplier is 8. All data-driven models are trained using Adam optimizer through 1000 epochs with an initial learning rate of 0.001 which is halved every 100 epochs.

    \subsection{Navier-Stokes Setup}
    The full PINNs for representative solutions of NS equation are trained through 40000 epochs with an initial learning rate of 0.005 halved after every 5000 epochs on a $100\times100$ grid with 100 time steps. The architecture of each PINN is $[3,20,20,20,20,20,20,2]$ which outputs both vorticity and stream field, and the activation function is $\cos$ and 5000 epochs are used during the offline and the online stage of fine tuning, with a fixed learning rate of 0.005, on the same mesh. 20 pre-trained neurons are developed for ReBaNO. In addition, PCA-Net is composed of four layers and 200 neurons per layer and DeepONet consists of a branch net with four layers of width 100 and a trunk net with three layers with the same width. FNO is composed of two Fourier neural layers with 20 neurons for each layer along with one lifting layer and one projection layer. The maximum Fourier modes used is set to be 6. CNO consists of four down/up sampling blocks, two residual blocks in the middle, and two residual blocks in the neck. The channel multiplier is 4. All data-driven models are trained using Adam optimizer through 1000 epochs with an initial learning rate of 0.001 which is halved every 100 epochs.

\subsection{Model size and computational cost}
\label{sec:suppl-figs}
\begin{table*}[htbp]
    \begin{small}
    \centering
    \caption{Size and computational cost of each model for Poisson (top), Darcy flow (middle), and Navier-Stokes (bottom).}
    \begin{tabular}{cccc}
        \toprule 
        model & \# of params & train time & infer time/$N_{\mathrm{test}}$ \\
        \midrule
        PCA-Net & \num{17670} & \qty{111.098}{\second} & \qty{0.163}{\ms} \\
        DeepONet & \num{19021} & \qty{132.788}{\second} & \qty{0.230}{\ms}\\
        FNO & \num{94657} &\qty{761.609}{\second} & \qty{1.235}{\ms} \\
        CNO & \num{103989} & \qty{504.188}{\second} & \qty{2.441}{\ms} \\
        PINO & \num{94657} & \qty{9808.844}{\second} & \qty{3.698}{\ms} \\
        ReBaNO & $8 + 901\times8$ & \qty{979.533}{\second} & \qty{0.399}{\second}  \\
        \bottomrule
        \midrule
        PCA-Net & \num{410733} & \qty{116.913}{\second} & \qty{0.132}{\ms} \\
        DeepONet & \num{384601} & \qty{179.098}{\second} & \qty{0.511}{\ms} \\
        FNO & \num{412929} & \qty{1105.024}{\second} & \qty{3.111}{\ms} \\
        CNO & \num{359877} & \qty{2672.568}{\second} & \qty{4.222}{\ms} \\
        ReBaNO & $48 + 8361\times48$ & \qty{6.328}{\hour} & \qty{1.940}{\second} \\
        \bottomrule
        \midrule
        PCA-Net & \num{86221} & \qty{94.039}{\second} & \qty{0.112}{\ms} \\
        DeepONet & \num{51601} & \qty{149.216}{\second} & \qty{0.189}{\ms} \\
        FNO & \num{58961} & \qty{287.531}{\second} & \qty{0.569}{\ms} \\ 
        CNO & \num{93783} & \qty{2232.984}{\second} & \qty{4.919}{\ms} \\
        ReBaNO & $20 + 2222\times20$ & \qty{40.774}{\hour} & \qty{14.092}{\second} \\
        \bottomrule
    \end{tabular}
    \label{tab:comp-cost}\\
    \vspace{6pt}
    *Training data generation time for the three equations are \qty{0.211}{\ms}/case, \qty{15.405}{\second}/case, and \qty{7.797}{\s}/case respectively. PINN training time for Poisson, Darcy flow, and Navier-Stokes are \qty{0.029}{\hour}/case, \qty{0.101}{\hour}/case, and \qty{1.952}{\hour}/case respectively.
\end{small}
\end{table*}

\end{document}